\documentclass[a4paper,fleqn]{cas-sc}
\usepackage{cite}
\usepackage[authoryear,round]{natbib}
\usepackage{setspace}
\usepackage{subfigure}

\usepackage{soul}
\usepackage{pifont}
\usepackage{color, xcolor}
\sethlcolor{yellow}
\usepackage{lineno}
\usepackage[ruled]{algorithm2e}

\def\tsc#1{\csdef{#1}{\textsc{\lowercase{#1}}\xspace}}
\tsc{WGM}
\tsc{QE}
\tsc{EP}
\tsc{PMS}
\tsc{BEC}
\tsc{DE}
\begin{document}
\soulregister{\citeyearpar}7
\soulregister{\citeyear}7
\soulregister{\ref}7
\soulregister{\citep}7
\soulregister{\cite}7
\let\WriteBookmarks\relax
\def\floatpagepagefraction{1}
\def\textpagefraction{.001}
\shorttitle{}

\shortauthors{Sun et al.}

\title [mode = title]{Multiple Prior Representation Learning for Self-Supervised Monocular Depth Estimation via Hybrid Transformer}          



\author[1,3]{Guodong~Sun}
\credit{}

\author[1,2,3]{Junjie~Liu}

\credit{}

\author[1,3]{Mingxuan~Liu}

\credit{}

\author[4]{Moyun~Liu}
\credit{}

\author[1,2,3,5]{Yang~Zhang}
\cormark[1] 
\ead{yzhangcst@hbut.edu.cn} 
\credit{}

\affiliation[1]{organization={School of Mechanical Engineering},
            addressline={Hubei University of Technology}, 
            city={Wuhan},
            postcode={430068}, 
            country={China}}
\affiliation[2]{organization={State Key Laboratory of Intelligent Optimized Manufacturing in Mining \& Metallurgy Process},
            addressline={ Beijing Key Laboratory of Process Automation in Mining \& Metallurgy}, 
            city={Beijing},
            postcode={102628}, 
            country={China}}
\affiliation[3]{organization={Hubei Key Laboratory of Modern Manufacturing Quality Engineering},
            addressline={Hubei University of Technology}, 
            city={Wuhan},
            postcode={430068}, 
            country={China}}
\affiliation[4]{organization={School of Mechanical Science and Engineering},
		  addressline={Huazhong University of Science and Technology}, 
			city={Wuhan},
			postcode={430074}, 
			country={China}}
\affiliation[5]{organization={National Key Laboratory for Novel Software Technology},
			addressline={Nanjing University}, 
			city={Nanjing},
			postcode={210023}, 
			country={China}}

\cortext[1]{Corresponding author: Yang~Zhang}


\begin{abstract}
Self-supervised monocular depth estimation aims to infer depth information without relying on labeled data. However, the lack of labeled information poses a significant challenge to the model's representation, limiting its ability to capture the intricate details of the scene accurately. Prior information can potentially mitigate this issue, enhancing the model's understanding of scene structure and texture. Nevertheless, solely relying on a single type of prior information often falls short when dealing with complex scenes, necessitating improvements in generalization performance. To address these challenges, we introduce a novel self-supervised monocular depth estimation model that leverages multiple priors to bolster representation capabilities across spatial, context, and semantic dimensions. Specifically, we employ a hybrid transformer and a lightweight pose network to obtain long-range spatial priors in the spatial dimension. Then, the context prior attention is designed to improve generalization, particularly in complex structures or untextured areas. In addition, semantic priors are introduced by leveraging semantic boundary loss, and semantic prior attention is supplemented, further refining the semantic features extracted by the decoder. Experiments on three diverse datasets demonstrate the effectiveness of the proposed model. It integrates multiple priors to comprehensively enhance the representation ability, improving the accuracy and reliability of depth estimation. Codes are available at: \url{https://github.com/MVME-HBUT/MPRLNet}.
\end{abstract}


\maketitle
\doublespacing

\section{Introduction}

Depth estimation is crucial in fields like autonomous driving and 3D scene understanding \citep{fields1}. Given the significant labeling costs, self-supervision has emerged as a cost-effective alternative, leveraging the inherent geometry of objects and spatial relationships within images to train models without relying on labeled data. Nevertheless, the absence of labeled information affects the representation ability of the model. Meanwhile, many untextured areas and complex structures are in the scenes, which challenges the model's generalization. To address these issues, incorporating prior information has been proposed to enhance model performance \citep{prior, prior1}. However, the information learned from a single prior is limited, and the effective complexity of the model is much lower than its representation ability \citep{complexity}. There remains a need to explore integrating scene information from diverse dimensions. By leveraging multiple prior information, we can exploit the potential of self-supervised monocular depth estimation, enabling a more accurate and robust understanding of 3D scenes.

Spatial priors enable the model to understand objects' motions and relative positions within a scene. To capitalize on this, vision transformer \citep{transformer1} splits the image into multiple patches and embeds them into the transformer encoder. However, as all patches are processed simultaneously, local detailed features are frequently ignored. Xu et al. \citeyearpar{Coat} addressed this by parallelizing cross-layer attention, thus representing both coarse and fine features. The drawback is the increase in computational cost. In addition, for the pose backbone, researchers employed common residual networks \citep{resnet, residual} to extract the pose relationships of image sequences. These networks are limited to local processing and lack long-range information. Hence, a lightweight and globally aware network is needed to address training challenges.

In the preceding analysis, we have discussed the acquisition of spatial priors. For the extracted information, Ronneberger et al. \citeyearpar{Unet} introduced skip connections to fuse encoder and decoder features, compensating for the information lost during the sampling process. While this approach facilitates the flow of information across different levels, the direct bypassing of features at other levels can hinder the model's ability to capture intricate pixel dependencies. Huang et al. \citeyearpar{ccnet} utilized criss-cross attention to extract the context information. The limitation is that the criss-cross attention exclusively focuses on the information of the spatial dimension. Since the interdependence between channels is ignored, the overall generalization performance of the model will be affected. Given these considerations, how to capture the context relationship more comprehensively is the current problem to be solved.

In the training process, the self-supervised approach integrates spatial pose information with depth information, realizing the reconstruction of target views and the subsequent calculation of reprojection loss \citep{monodepth2}. To solve the problem of image blur in prediction, Godard et al. \citeyearpar{monodepth} calculated smoothness loss between the input images and the depth maps. However, photometric differences often lead to the loss of landscapes, hampering the optimization process. Shu et al. \citeyearpar{featureloss} proposed a feature metric loss, focusing on the feature representation to constrain the loss landscape. Despite these advancements, depth maps still struggle with weak textures and ambiguous object boundaries, resulting in boundary scale deviation. As a dense prediction task, semantic segmentation classifies the objects at the pixel level and splits the outlines. Combining semantic priors with depth estimation is key to solving the boundary scaling issue.

\begin{figure}[t]
	\centering
		\includegraphics[scale=.6]{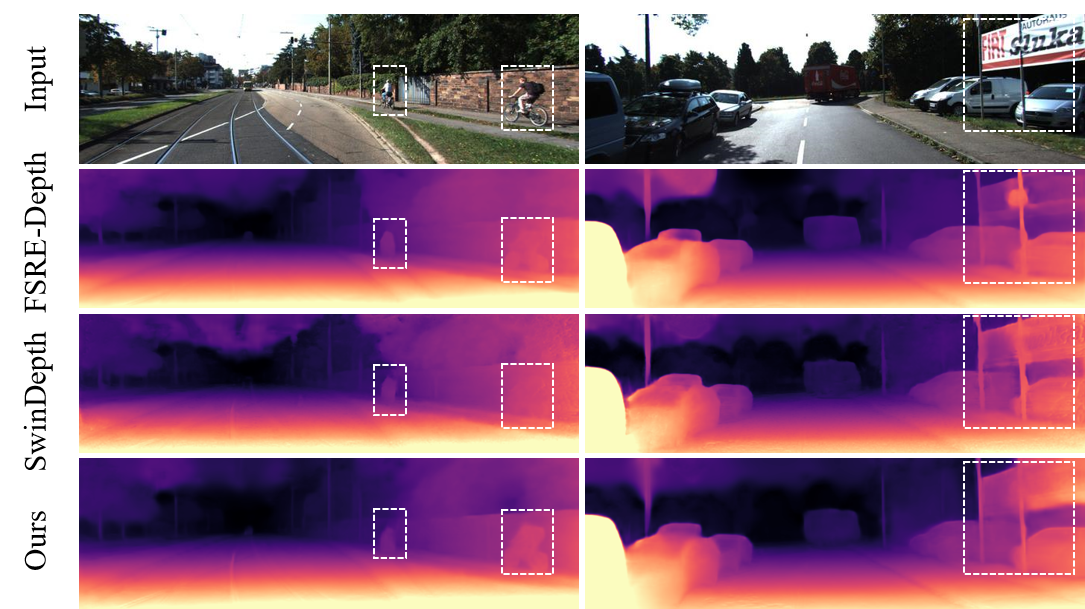}
	\caption{Comparison of depth maps. Benefiting from learning multiple priors, the proposed network captures local details and correct contours. In terms of representation, our method is superior to FSRE-Depth \citep{fsre} and SwinDepth \citep{Swindepth}. For example, it estimates clearer contours of cyclists and billboards.}
	\label{FIG:1}
\end{figure}

To address these challenges, we introduce a novel self-supervised monocular depth estimation network guided by multiple priors. In our approach, the depth encoder adopts a hybrid transformer architecture, embedding multi-scale patches in the transformer encoder layer to obtain spatial prior information. Additionally, a long-range attention mechanism and low-cost operation are used in the pose encoder. To further enhance the representation, we introduce a context prior attention mechanism that processes encoder features, capturing context priors across different pixels. For semantic priors, we adopt semantic images as semantic pseudo-labels and use semantic boundary loss to guide the model for training. In addition, we redesign the decoder structure and apply neuroscience theory to strengthen the semantic representation. Figure \ref{FIG:1} illustrates the depth maps obtained from different networks \citep{fsre, Swindepth}, where our model generates more complete contours and clearer details. The experiment on the KITTI dataset evaluates the effectiveness of multiple priors, and the proposed method outperforms existing methods, with 0.104 in $\text{Abs Rel}$ and 0.705 in $\text{Sq Rel}$. Then, the generalization performance is demonstrated on the Make3D and NYU Depth V2 datasets. In ablation experiments, we verify the rationality of the proposed modules. Overall, the main contributions are as follows:

\begin{itemize} \item Leveraging a hybrid transformer and a lightweight pose network to model long-range dependence in the spatial dimension enhances the model's ability to capture global spatial relationships. \item We introduce a context prior attention (CPA) mechanism specifically tailored to perceive surrounding pixels in regions with complex structures or limited texture, which significantly enhances the generalization capability. \item The semantic priors are introduced through the semantic boundary loss (SBL) to solve the boundary scale deviation, and the semantic prior attention (SPA) mechanism is supplemented to refine the semantic features. \item We explore complementary cues of spatial, context, and semantic prior information to enhance representation learning capability for self-supervised monocular depth estimation. The model has advanced performance and generalization.
\end{itemize}

The remainder of the article is organized as follows: Section II introduces related works. Section III details our proposed method. Section IV contains comparative experiments on the three datasets, ablation experiments, and discussion. Section V concludes the paper.



\section{Related work}

\subsection{Self-supervised monocular depth estimation}
Self-supervised monocular depth estimation has garnered significant attention in recent years due to its potential for reducing training costs. Zhou et al. \citeyearpar{ego-motion} pioneered using unlabeled image sequences for training, leveraging a pose network to capture the camera's ego-motion and combining it with depth information to reconstruct target views. This approach laid the foundation for subsequent advances in the field. Godard et al. \citeyearpar{monodepth2} further utilized automatic masks to ignore relatively stationary pixels and improved robustness with minimum reprojection loss. Despite improved training methods, the uncertainty of estimating depth maps without ground truth has not been explored. Poggi et al. \citeyearpar{mono-uncertainty} introduced how to deal with uncertainty modeling and proposed uncertainty estimation by image flipping, empirical estimation, predictive estimation, and Bayesian estimation. Previous methods dealt with independent data frames, but continuous depth information requires updating in practice. Patil et al. \citeyearpar{Patiletal} exploited the spatio-temporal structure in the data frames to obtain more accurate depth information. Wang et al. \citeyearpar{Wangetal} utilized dynamic and static cues to generate helpful depth recommendations. The structure and cues of data have driven the practical applications of depth estimation. To further focus on regions of interest within the scene, Yan et al. \citeyearpar{caddepth} employed channel attention to recalibrate channel-wise features and emphasize semantic information. Song et al. \citeyearpar{MLDA-Net} combined global and local attention to reinforce structural information. Johnston et al. \citeyearpar{johnston} utilized self-attention to explore context information, which can infer similar disparity values in discontinuous regions. The employment of attention mechanisms boosted the scene perception capabilities of depth estimation. Furthermore, considering that information in complex scenes varies in scale, Zhou et al. \citeyearpar{r-msfm} adopted multi-scale feature modulation to achieve a more refined representation and richer semantics. Zhang et al. \citeyearpar{DynaDepth} learned the absolute scale during training and had better robustness. Wang et al. \citeyearpar{ScaleInvariant} achieved scale consistency through point cloud alignment to enhance self-supervision training.
However, due to the lack of sufficient supervision signals, the improvement of the framework is constrained. Zhou et al. \citeyearpar{VC-Depth} introduced velocity constancy as a supplementary supervision signal. Chawla et al. \citeyearpar{G2S} suggested a dynamically weighted GPS-to-scale loss to offset the appearance-based loss. Lee et al. \citeyearpar{Leeetal} combined instance-aware photometry and photometric consistency for self-supervision. These efforts have expanded the directions for overcoming the limitations of self-supervised monocular depth estimation frameworks.

The aforementioned works have demonstrated the diversity and development of self-supervised monocular depth estimation methods. Researchers have continuously improved the depth estimation performance by introducing techniques such as automatic masks, temporal-spatial structures, attention mechanisms, and additional supervised signals. Although these methods have succeeded in certain areas, they still face challenges and limitations. For instance, when confronted with varying environments, objects, and scene layouts, the performance of the models often deteriorates. This is primarily due to the models' over-reliance on specific data distributions and feature representations during training. Additionally, under different lighting conditions, the shadows on the surfaces of objects change, affecting the extraction of depth information. Therefore, further exploring effective information within scenes to enhance the stability and generalization capability of depth estimation remains an important direction for future research.
\subsection{Prior information}
In image processing, prior information refers to previous experience or known information that can assist the algorithm in understanding the scene. Ulyanov et al. \citeyearpar{deepimageprior} proposed that prior knowledge can be extracted from a few images to help solve image inpainting problems, such as denoising, super-resolution, restoration, etc. With the development of deep learning, vision transformers \citep{transformer} have become powerful tools for extracting dense features. Still, their convergence speed is slow due to the lack of spatial induction bias. Zhou et al. \citeyearpar{spvit} addressed this limitation by introducing spatial priors, which generalize locally constrained convolutional inductive biases to local and nonlocal correlations. This advancement enables learning beneficial 2D spatial inductive biases, enhancing the transformers' ability to interpret spatial relationships within images. Chen et al. \citeyearpar{spatialadapter} further refined this concept by introducing spatial prior into the vision transformer to reorganize the fine-grained multi-scale features. Moreover, Yu et al. \citeyearpar{contextprior} proposed a different approach, applying context prior to selectively capturing intra-class and inter-class context correlation to achieve reliable feature representation. Wang et al. \citeyearpar{nonlocal} adopted self-attention to extract the relationship between a single feature and features at all other positions to obtain context information. Semantic information has also been explored as a prior for training \citep{Guizilini, SAFENet}. Jung et al. \citeyearpar{fsre} integrated a semantic decoder into the depth estimation network and then refined the depth features in the metric learning formulation by using semantic prior information. Klingner et al. \citeyearpar{SGdepth} proposed combining depth estimation and semantic segmentation for cross-domain training. By exploiting the complementarity of the two tasks, they have achieved mutually beneficial results. In addition, to deal with fuzzy and uncertain information, Zia et al. \citeyearpar{zia} proposed a complex linear Diophantine fuzzy set to handle ambiguity with more efficient responses. Zhao et al. \citeyearpar{zhao} extended fuzzy rough clustering to the algorithm and explored cluster medoids from multiple perspectives.

In summary, depth estimation requires perceiving objects of different scales in complex scenes and capturing the relative relationships between each pixel. Introducing prior information can enhance the algorithm's perception and understanding capabilities. As the demand for depth estimation becomes increasingly complex, relying solely on a single type of prior information often fails to achieve the desired processing effect. Therefore, combining multiple types of prior information has become a research direction worth exploring. Spatial priors help to understand the structure and layout of images. However, a single spatial prior lacks context information. Semantic priors provide more precise scale information, but adding a semantic decoder can increase computational complexity. Based on the analysis of existing depth estimation methods, we propose a framework to explore further the complementary cues of spatial, context, and semantic priors for monocular depth estimation.
\section{Method}
\begin{algorithm}[!t]
    \caption{Overall process of the method} \label{algorithm1}
    Framework: DepthNet, PoseNet\\
Loss function: Semantic boundary loss, Reprojection loss, Smoothness loss\\
    \eIf{Train}{
    \KwIn{Image sequence ($I_{0}$, $I_{1}$), Semantic pseudo-labels ($S$);}
    $I_{0}$ $\to$ DepthNet $\Rightarrow$ depth information ($D$);\\
    $I_{0}$, $I_{1}$ $\to$ PoseNet $\Rightarrow$ camera pose ($P$);\\
    $I_{1}$, $D$, $P$ $\to$ Reprojection $\Rightarrow$ $I_{1}^{'}$;\\
    $I_{1}^{'}$, $I_{0}$ $\to$ Reprojection loss;\\
    $D$, $I_{0}$ $\to$ Smoothness loss;\\
    $D$, $S$ $\to$ Semantic boundary loss;\\
    Backpropagation and parameter updates.
            }
        {
         Perform inference\\
         \KwIn{Image ($I$);}
         $I$ $\to$ DepthNet $\Rightarrow$ sigmoid output;\\
         Convert sigmoid output into depth map;\\
         \KwOut{depth map.}
        } 
\end{algorithm}
In this section, we first introduce the overview of our motivation. Following that, we explore spatially augmented encoders to perceive scene structure. Subsequently, context prior attention is presented to extract the context prior from the encoder features. Finally, we introduce semantic prior pseudo-labels to calculate semantic boundary loss and propose the semantic prior layer.
\subsection{Motivation overview}
Perceiving the spatial structure of the scene is the basis for monocular depth estimation. Due to the limitation of the receptive field, it is difficult for convolution-based methods \citep{monodepth2, caddepth} to distinguish between different objects. Simultaneously, depth information cannot be accurately inferred only by a pixel or object. If some pixels have shadows, it will affect the accuracy of the model. Since the lack of semantic guidance, the depth regions often exceed the actual contour for complex scenes and weak texture regions, called boundary scale deviation. Given the above considerations, we introduce spatial priors, context priors, and semantic priors into the depth estimation task to enhance the representation and generalization.

\begin{figure}[!t]
	\centering
		\includegraphics[width=6.3in]{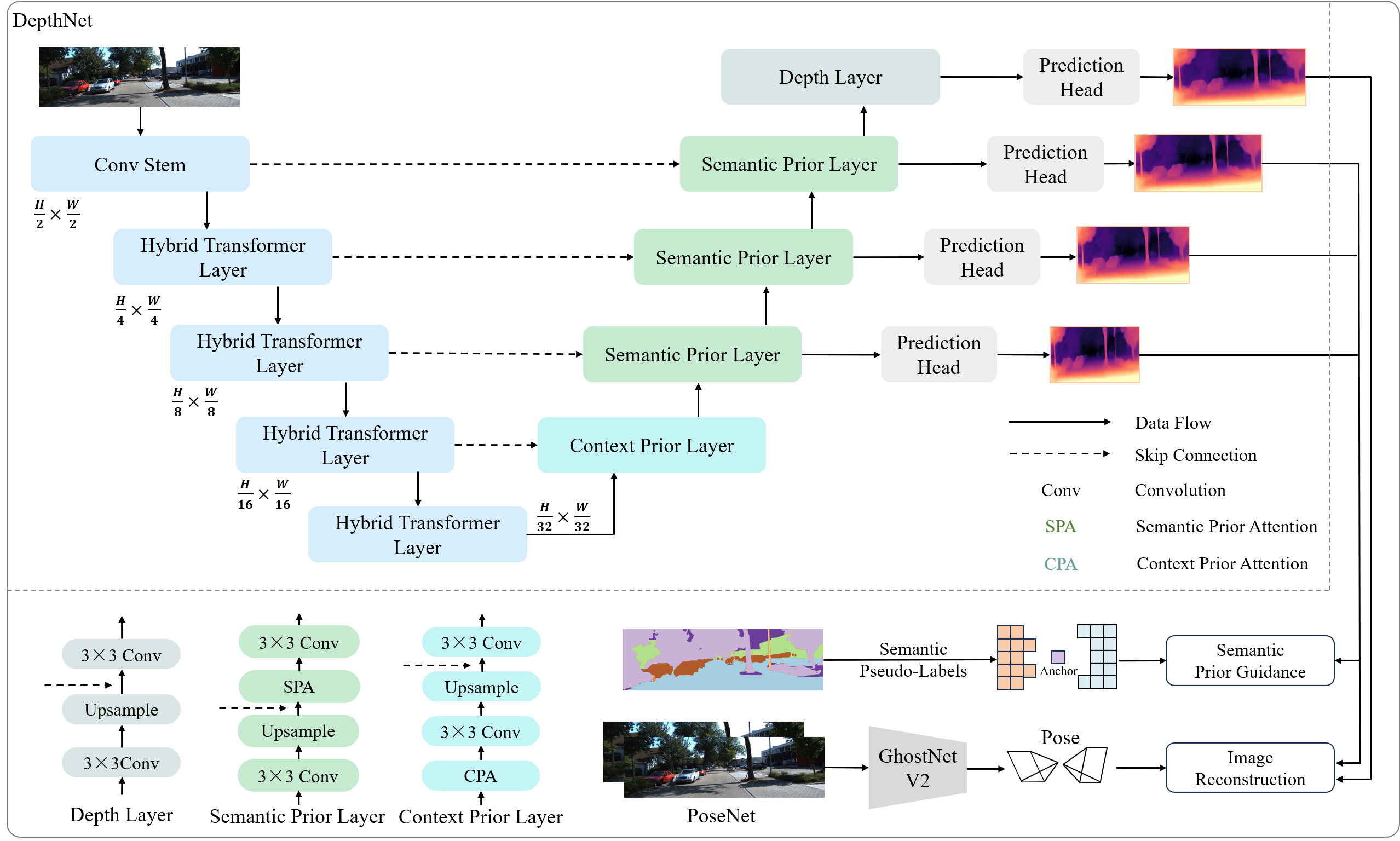}
	\caption{Overview of the proposed model. The model employs a spatially augmented depth encoder and a lightweight PoseNet to extract spatial priors. The context prior layer in the decoder is used to capture the context priors. Meanwhile, semantic pseudo-labels are introduced into the semantic boundary loss, supplemented by the semantic prior layer to provide semantic prior guidance. The depth decoder consists of a context prior layer, three semantic prior layers, and a depth layer, where the context prior layer and semantic prior layer contain CPA and SPA, respectively. Note that only DepthNet works for inference.}
	\label{Fig. 2}
\end{figure}

As shown in Fig. \ref{Fig. 2}, we design a self-supervised monocular depth estimation model guided by multiple priors. In DepthNet, the hybrid transformer encoder extracts the spatial priors. The first layer of the decoder applies context prior attention (CPA), and the second to fourth layers contain semantic prior attention (SPA) as semantic prior layers. In PoseNet, we employ a lightweight encoder based on the long-range attention mechanism to extract spatial pose priors. Furthermore, the semantic prior pseudo-labels are introduced into the network to guide the semantic prior layers for training through semantic boundary loss (SBL).

In Algorithm \ref{algorithm1}, we present the overall process of the method. The model comprises a DepthNet and a PoseNet. During training, the input is adjacent image sequence ($I_{0}$, $I_{1}$) and semantic pseudo-labels ($S$). The depth network extracts depth information ($D$) from $I_{0}$, while the pose network determines the camera pose ($P$) between $I_{0}$ and $I_{1}$. Then, $I_{1}$ is reprojected through $D$ and $P$ to obtain the reconstructed image $I_{1}^{'}$ in view of $I_{0}$. Subsequently, we calculate the reprojection loss between $I_{1}^{'}$ and $I_{0}$, the smoothness loss between $D$ and $I_{0}$, and the semantic boundary loss between $D$ and $S$. Then, backpropagation and parameter updates are performed. During inference, the input is image $I$. The model only utilizes the DepthNet to generate sigmoid output and further converts the sigmoid output into the depth map.

\subsection{Spatial prior}
Most methods adopt convolutional backbones \citep{monodepth2, caddepth}, which are limited to local regions and usually ignore some important spatial global information. Inspired by MPViT \citep{mpvit}, we utilize the transformer architecture to understand the scenes' spatial structure. The backbone has five stages as shown in Fig. \ref{Fig. 2}. The first stage is the conv stem, followed by four hybrid transformer layers.

\begin{figure}[!t]
	\centering
		\includegraphics[scale=.35]{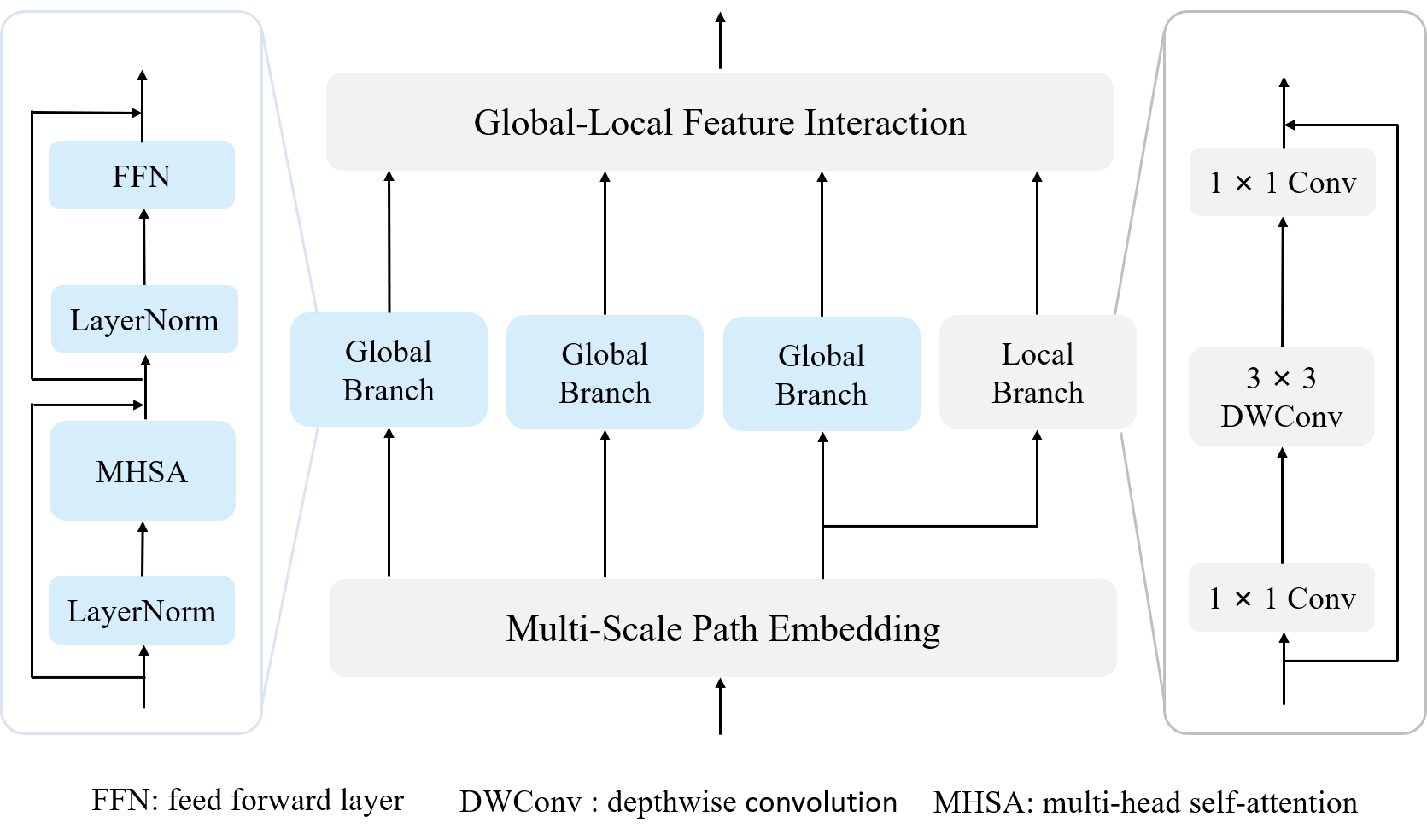}
	\caption{Structure of the hybrid transformer layer. Through multi-scale path embedding, receptive fields of different scales can be obtained. Then, the transformer and convolution branches capture global spatial features and local detail features. Subsequently, the global-local feature interaction layer fuses these features.}
	\label{Fig. 3}
\end{figure}

The conv stem employs two $3\times 3$ convolutions where the image is downsampled to generate features of size ${\frac{H}{2}}\times{\frac{W}{2}}$. From the second to the fifth stage, each hybrid transformer layer contains a multi-scale patch embedding, as shown in Fig. \ref{Fig. 3}. Convolutional layers with different kernel sizes have varying receptive fields, perceiving objects or regions of different scales. To reduce the number of parameters, stacking consecutive convolutions of the same filter size can also expand the receptive field \citep{monovit}, such as two $3\times3$ are equivalent to $5\times5$. We employ three parallel convolution layers in the multi-scale patch embedding, each with $3\times3$ depthwise convolutions and pointwise convolution followed by batch normalization and HardSwish activation. Through the above operations, multi-scale embedding tokens are obtained. Since the transformer is conducive to extracting global information while local details are often ignored, the extracted tokens are dealt with parallel and complementary by the global branches and the local branch. The second stage has a local branch and two global branches, while the number of global branches in the third to fifth stages becomes three. In the global branch, we project image tokens to query ($\mathbf{Q}$), key ($\mathbf{K}$), and value ($\mathbf{V}$), and use factorized self-attention \citep{Coat} mechanism to capture inter-pixel relationships $\mathbf{G}$:
\begin{equation}
\text{FactorAtt}(\mathbf{Q},\mathbf{K},\mathbf{V})=\frac{\mathbf{Q}}{\sqrt{C}}(\text{softmax}(\mathbf{K})^T\mathbf{V}),
\end{equation}
where $C$ refers to the embedding dimension. Moreover, to extract local features, we adopt a depthwise bottleneck structure in the local branch, which consists of $1\times1$ convolution, $3\times3$ depthwise convolution, and $1\times1$ convolution. Finally, we adopt a global-local feature interaction layer to fuse local features and global features.

Through the above depth encoder, we perceive the spatial structure of the scene. Further, to obtain the spatial pose priors of the image sequences, we adopt the lightweight GhostNetV2 \citep{ghostnetv2} as the pose encoder. Specifically, cheap operations are adopted to reduce the computational cost. Half of the features are captured by $3\times3$ depthwise convolution, and the remaining features are obtained by $1\times1$ pointwise convolution. Meanwhile, to capture the long-range spatial prior information of adjacent frames, aggregating features in vertical and horizontal directions with decoupled fully connected attention \citep{ghostnetv2}. Compared with the ordinary fully connected layer \citep{fullyconnected}, this operation enhances the spatial expressiveness of the model and has lower computational complexity.

\subsection{Context prior}
\begin{figure}[!t]
	\centering
		\includegraphics[scale=.4]{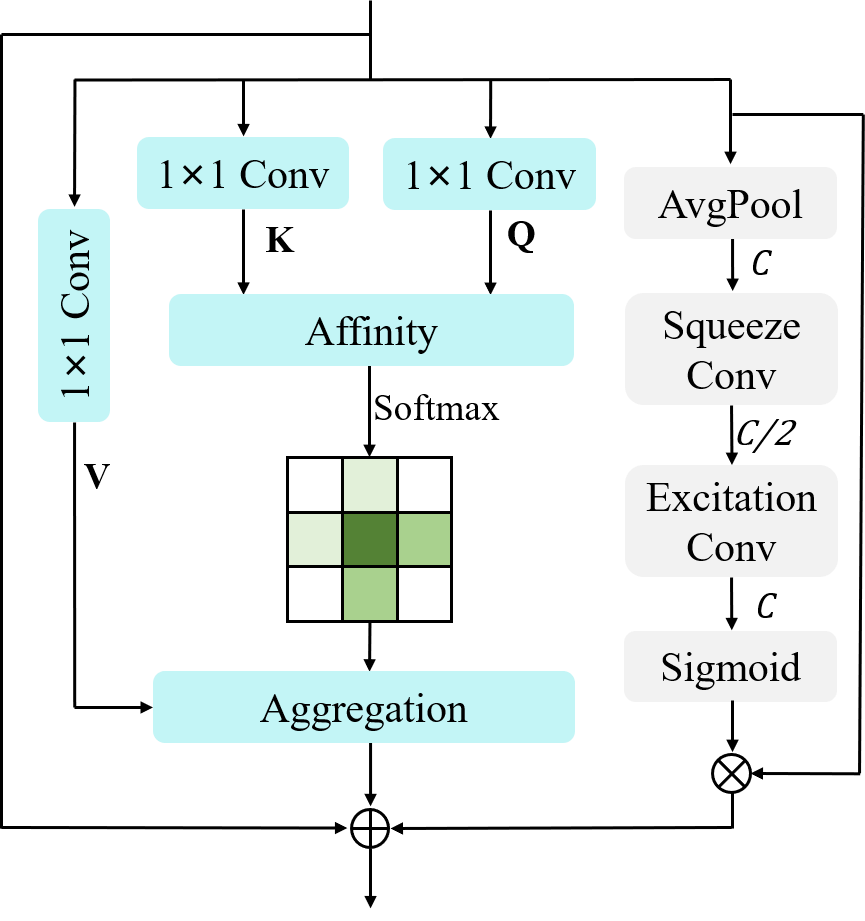}
	\caption{The structure of the proposed context prior attention. CPA uses the spatial branch (cyan) to extract the context information in the spatial dimension while supplementing the relationship in the channel dimension with the channel branch (gray). The $\mathbf{Q}$, $\mathbf{K}$, $\mathbf{V}$ represent the query, key, and value.}
	\label{Fig. 4}
\end{figure}

When facing complex scenes and textureless regions, the generalization of depth estimation is challenged. Wang et al. \citeyearpar{nonlocal} adopted a self-attention mechanism to extract the relationship between each pixel. Huang et al. \citeyearpar{ccnet} improved the efficiency by using criss-cross attention to obtain the context relationship between the cross-paths and surrounding pixels. Despite this, the above method only interests the spatial dimension and lacks the representation of channel information. We propose a context prior attention, detailed in Fig. \ref{Fig. 4}.

The CPA consists of a spatial branch (cyan) and a channel branch (gray). Given an input feature  $\mathbf{H}\in\mathbb{R}^{C\times W\times H}$, in the spatial branch, the input feature is processed through three $1\times1$ convolutions to generate $\mathbf{Q}$, $\mathbf{K}$, and $\mathbf{V}$. Specifically, $\{\mathbf{Q},\mathbf{K}\}\in\mathbb{R}^{C^{\prime}\times W\times H}$. $C^{\prime}$ represents the number of channels, which is $1/8$ of $C$. Meanwhile, $\mathbf{V}\in\mathbb{R}^{C\times W\times H}$. Subsequently, we generate an attention map $\mathbf{T}\in\mathbb{R}^{(H+W-1)\times(W\times H)}$ on $\mathbf{Q}$ and $\mathbf{K}$ by affinity operation. At a spatial position $\mathbf{u}$ in $\mathbf{Q}$, we can obtain the vector $\mathbf{Q_{u}}\in\mathbb{R}^{C^{\prime}}$. Also, in $\mathbf{K}$, we extract the set $\boldsymbol{\Omega_{u}}\in\mathbb{R}^{(H+W-1)\times C^{\prime}}$, which represents the vectors in the same row or column as $\mathbf{u}$. $\boldsymbol{\Omega}_{i,\mathbf{u}}\in\mathbb{R}^{C^{\prime}}$ is the $i-th$ vector in $\mathbf{\Omega}_{\mathbf{u}}$. The affinity operation is defined as 
$p_{i,\mathbf{u}}=\mathbf{Q_u}\Omega_{i,\mathbf{u}}^\intercal$,
where $p_{i,\mathbf{u}}\in\mathbf{P}$ represents the degree of correlation between $\mathbf{Q_{u}}$ and $\boldsymbol{\Omega}_{i,\mathbf{u}}$. The $\begin{aligned}\mathbf{P}&\in\mathbb{R}^{(H+W-1)\times(W\times H)}\end{aligned}$ and $i=[1,...,H~+~W~-~1]$. Then, we calculate attention map $\mathbf{T}$ through the softmax operation.
At each spatial position in $\mathbf{V}$ corresponding to $\mathbf{u}$, we can obtain the vector $\mathbf{V}_{\mathbf{u}}\in\mathbb{R}^{C}$. The vectors in the same row or column as $\mathbf{V}_{\mathbf{u}}$ form the set $\Phi_{\mathbf{u}}\in\mathbb{R}^{(H+W-1)\times C}$. We extract the context information of the spatial dimension through the aggregation operation:
\begin{equation}
\mathbf{H}_{\mathbf{u}}^{\prime}=\sum_{i=0}^{H+W-1}\mathbf{T}_{i,\mathbf{u}}\mathbf{\Phi}_{\mathbf{i,u}},
\end{equation}
where $\mathbf{H_u^{\prime}}$ is the feature vector of position $\mathbf{u}$ in $\mathbf{H^{\prime}}\in\mathbb{R}^{C\times W\times H}$. And $\mathbf{T}_{i,\mathbf{u}}$ is the scalar value located at position $\mathbf{u}$ and channel $i$ in $\mathbf{T}$.

\begin{figure}[!t]
	\centering
	\begin{minipage}[t]{0.24\linewidth}
		\vspace{3pt}
		\centerline{\includegraphics[width=\textwidth]{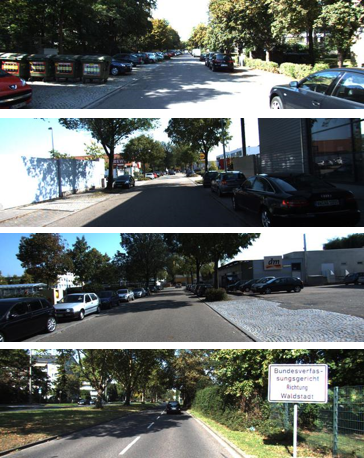}}
		\centerline{Input}
	\end{minipage}
	\begin{minipage}[t]{0.24\linewidth}
		\vspace{3pt}
		\centerline{\includegraphics[width=\textwidth]{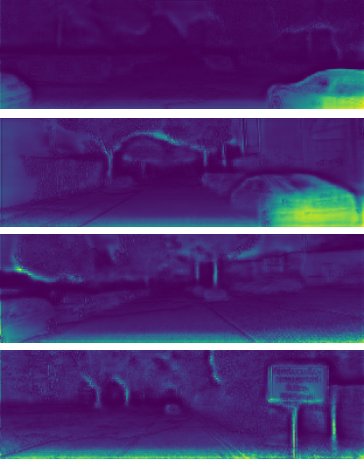}}
		\centerline{W/O attention}
	\end{minipage}
	\begin{minipage}[t]{0.24\linewidth}
		\vspace{3pt}
		\centerline{\includegraphics[width=\textwidth]{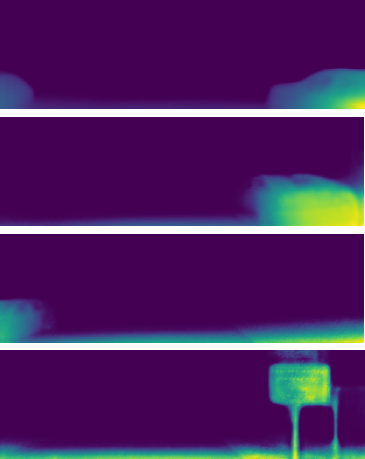}}
		\centerline{Criss-cross attention}
	\end{minipage}
	\begin{minipage}[t]{0.24\linewidth}
		\vspace{3pt}
		\centerline{\includegraphics[width=\textwidth]{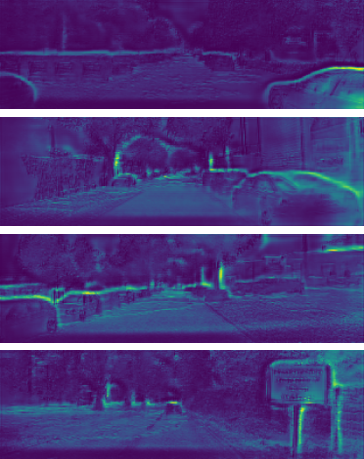}}
		\centerline{CPA}
	\end{minipage}
	\caption{Visualization of the attention mechanism. We list the visual features of CPA, criss-cross attention \citep{ccnet}, and the case without attention. The criss-cross attention focuses excessively on close objects. In contrast, our CPA focuses on nearby scenes while highlighting distant objects such as vehicles and trees.}
	\label{Fig. 5}
\end{figure}

In the channel branch, we treat the input feature $\mathbf{H}=[\mathbf{h}_1,\mathbf{h}_2,\cdots,\mathbf{h}_C]$, and $\mathbf{h}_{k}$ represents a channel. To obtain the relationship in channel dimension, global average pooling is used to generate vector $\mathbf{z}\in\mathbb{R}^{1\times1\times C}$, where the $k^{th}$ element is
$\mathbf{z}_k=\frac{1}{H\times W}\sum_{i}^{H}\sum_{j}^{W}\mathbf{h}_k(i,j)$.
Additionally, the vector $\mathbf{z}$ is transformed to $\hat{\mathbf{z}}=\mathbf{W}_{1}(\mathbf{W}_{2}\mathbf{z})$, in which $\mathbf{W}_{1}\in\mathbb{R}^{C\times\frac{C}{2}}$ and $\mathbf{W}_2\in\mathbb{R}^{\frac C2\times C}$ represent the weights of two $1\times1$ convolutional layers. Then, through a sigmoid layer $\sigma(\hat{\mathbf{z}})$, the dynamic range of $\hat{\mathbf{z}}$ changes to interval $[0,1]$. Subsequently, the vector $\hat{\mathbf{z}}$ is used to activate the input feature $\mathbf{H}$:
\begin{equation}
\mathbf{U}^{\prime}=[\sigma(\hat{\mathbf{z}_{1}})\mathbf{h}_{1},\sigma(\hat{\mathbf{z}_{2}})\mathbf{h}_{2},\cdots,\sigma(\hat{\mathbf{z}_{C}})\mathbf{h}_{C}].
\end{equation}
Finally, we aggregate the spatially enhanced feature $\mathbf{H}^{\prime}$, the channel attention feature $\mathbf{U}^{\prime}$, and the original feature $\mathbf{H}$:
\begin{equation}
\mathbf{E}=G(\mathbf{H}^{\prime}+\mathbf{U}^{\prime})+\mathbf{H},
\end{equation}
in which $G$ is a trainable parameter and the enhanced context prior feature $\mathbf{E}$ is obtained. In addition, Figure \ref{Fig. 5} provides visual features of CPA, criss-cross attention \citep{ccnet}, and the case without attention. When the attention mechanism is not used, the perception of distant objects is mainly focused on trees, and vehicles are ignored. At the same time, the criss-cross attention focuses excessively on close objects. In contrast, CPA focuses on near scenes while highlighting distant vehicles and trees.

\subsection{Semantic prior}
To accurately distinguish different objects and solve the problem of boundary scale deviation. We design the semantic prior layer to enhance the semantic representation of the depth decoder and introduce the semantic priors through the semantic boundary loss.

\begin{figure}[!t]
	\centering
		\includegraphics[scale=.4]{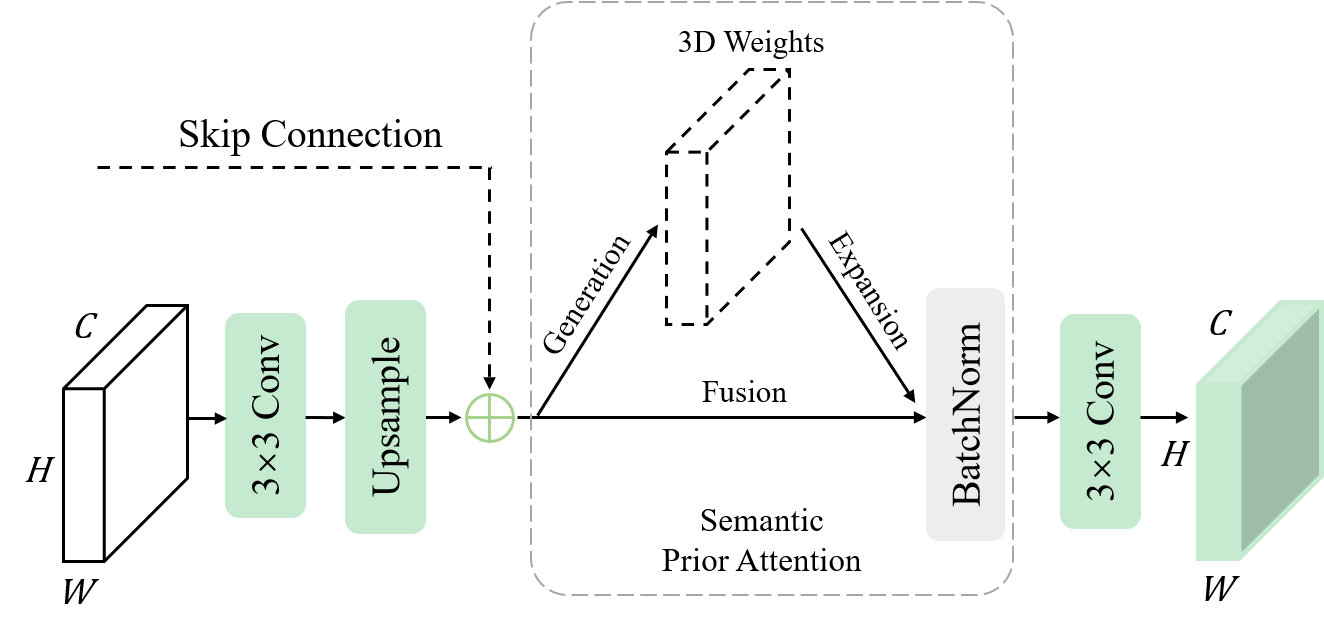}
	\caption{Details of the semantic prior layer. The layer convolves and upsamples the input features, then concatenates the upsampled features with skip connection features. Subsequently, features are enhanced through semantic prior attention, followed by a convolutional layer.}
	\label{Fig .6}
\end{figure}

Figure \ref{Fig .6} depicts the semantic prior layer, which convolves and upsamples the features and then concatenates the upsampled features with the skip connection features. Subsequently, the features are enhanced through the semantic prior attention, followed by a convolutional layer. The SPA is based on neuroscience theories, in which informative neurons tend to have different firing patterns from surrounding neurons, and surrounding neurons can be inhibited by active neurons \citep{neuron}. In general, active neurons have more important information and should be prioritized. Similar to \citep{simam}, we search for active neurons by measuring the linear separability of target neurons from other neurons. First, each neuron is defined as the following energy function:
\begin{equation}\label{equ:energy}
f_t(w_t,b_t,\mathbf{y},x_i)=(y_t-\hat{t})^2+\frac{1}{K-1}\sum_{i=1}^{K-1}(y_o-\hat{x}_i)^2.
\end{equation}

In the formula, $\hat{t}=w_tt+b_t$ and $\hat{x}_i=w_tx_i+b_t$, while $t$ and $x_{i}$ represent the target neuron and other neurons in a single channel, respectively. $K=H\times W$ is the number of neurons in the channel, and $i$ is the index in the spatial dimension. $\omega_{t}$ and $b_{t}$ are weight and bias the transform. Through Eq.~(\ref{equ:energy}), we can obtain the linear separability between the target neuron $t$ and other neurons in the same channel. To simplify the process, we use binary labels $(1,-1)$ for $y_{t}$ and $y_{0}$, and add a regularizer to the equation: 
\begin{equation}\label{equ:energy1}
f_{t}(w_{t},b_{t},\mathbf{y},x_{i}) =\frac1{K-1}\sum_{i=1}^{K-1}(-1-(w_tx_i+b_t))^2+(1-(w_tt+b_t))^2+\rho w_t^2.
\end{equation}
Then we can solve Eq.~($\ref{equ:energy1}$) through a closed-form solution with $w_{t}$ and $b_{t}$, which are contained by:
\begin{equation}
w_t=-\frac{2(t-\mu_t)}{(t-\mu_t)^2+2\sigma_t^2+2\rho}, \quad
b_t=-\frac12(t+\mu_t)w_t.
\end{equation}
The $\mu_t=\frac1{K-1}\sum_{i=1}^{K-1}x_i$ and $\sigma_t^2=\frac{1}{K-1}\textstyle \sum_{i}^{K-1}(x_i-\mu_t)^2$ are the mean and variance calculated by all neurons except $t$ in the channel. To reduce the computational cost, we assume that all pixels in a channel follow the same distribution. Thus, we can calculate the mean and variance of all neurons and apply them to the neurons in the channel. With the following formula, we can calculate the minimum energy:
\begin{equation}\label{equation et}
f_t^*=\frac{4(\hat{\sigma}^2+\rho)}{(t-\hat{\mu})^2+2\hat{\sigma}^2+2\rho},
\end{equation}
where $\hat{\mu}=\frac{1}{K}\textstyle\sum_{i=1}^{K}x_i$, $\hat{\sigma}^2=\frac{1}{K}\textstyle\sum_{i=1}^K(x_i-\hat{\mu})^2$, and the $\rho$ is set to $1\times10^{-4}$. The lower the $f_{t}^{*}$ is, the more different the neuron $t$ is from its surrounding neurons. Therefore, we adopt $1/f_t^*$ to represent the importance of each neuron and then refine it by 
$\widetilde{\mathbf{H}}=\mathbf{B}(sigmoid(\frac1{\mathbf{F}})\odot\mathbf{H})$,
where $\mathbf{F}$ groups all $f_t^*$ across channel and spatial dimensions, $sigmoid$ can prevent excessive value in $\mathbf{F}$. The $\mathbf{B}$ represents batch normalization to improve the robustness.
\begin{figure}[!t]
	\centering
		\includegraphics[scale=.45]{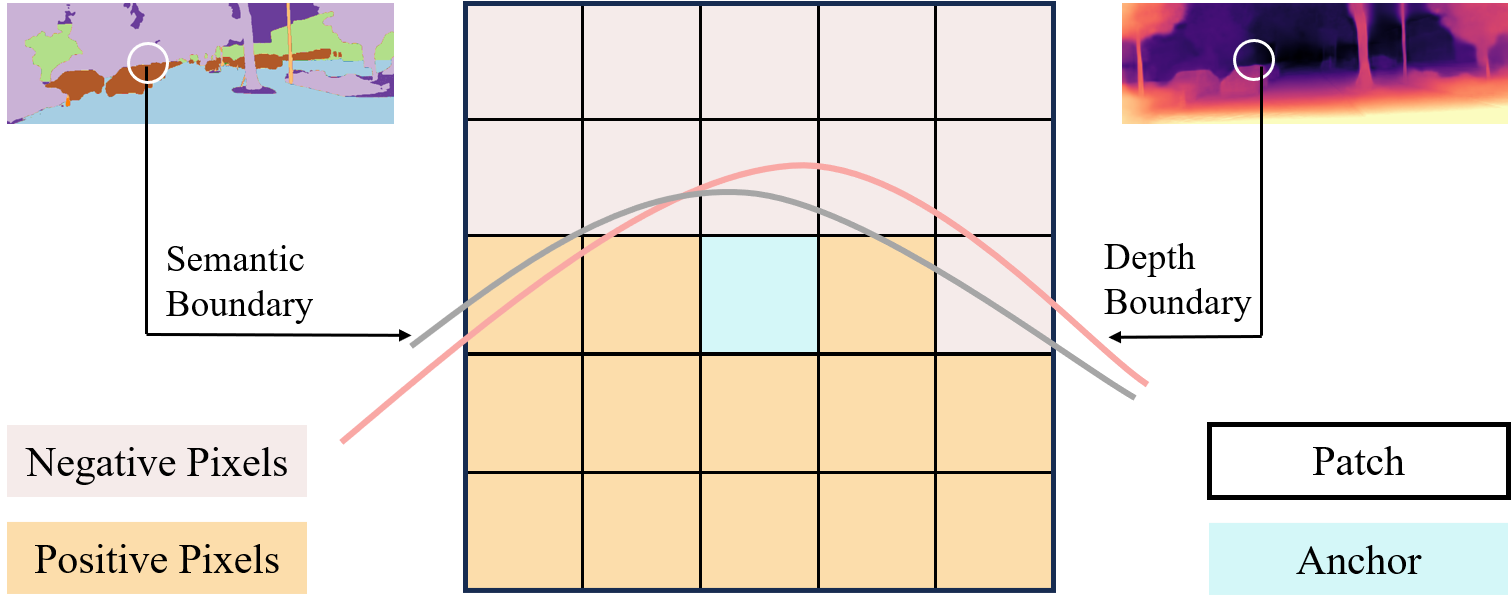}
	\caption{Details of the semantic boundary loss. We divide the semantic map into patches of $5 \times 5$ size, and the center of the patch is the anchor point. Pixels with the same semantic category as the anchor are positive, and others are negative.}
	\label{Fig .7}
\end{figure}

During training, we further introduce semantic priors into the model. Specifically, we generate semantic pseudo-labels with the pre-trained semantic segmentation model \citep{label} and then calculate the semantic boundary loss between the semantic pseudo-labels and the depth maps of the semantic prior layers. Figure \ref{Fig .7} depicts the semantic boundary loss. The pixel depth of the same object is generally similar, while the depth of different objects across the object boundary varies greatly \citep{tridepth}. We divide the semantic map into patches of $5 \times 5$ size and take the center pixel as the anchor. The position of the anchor is $i$. Furthermore, pixels in the same semantic category as the anchor are positive, and pixels in different semantic categories are negative. We define the sets of positive and negative pixels in the patch $\mathcal{P}_{i}$ as $\mathcal{P}_{i}^{+}$ and $\mathcal{P}_{i}^{-}$, respectively.
 The $\mathcal{P}_{i}^{+}$ and $\mathcal{P}_{i}^{-}$ can indicate whether the patch crosses the semantic boundary. When both $\mathcal{P}_{i}^{+}$ and $\mathcal{P}_{i}^{-}$ are larger than 0, $\mathcal{P}_{i}$ crosses the boundary. In the depth feature map, we classify the patches according to the corresponding pixel category in the semantic map. The positive pixel distance $d^{+}$ and the negative pixel distance $d^{-}$ are defined as the mean of Euclidean distance of $L_{2}$ normalized depth features:
\begin{equation}
d^+\left(i\right)=\frac1{\left|\mathcal{P}_i^+\right|}\sum_{j\in\mathcal{P}_i^+}\left \|\hat{F}_d\left(i\right)-\hat{F}_d\left(j\right)\right \|_2^2,
\quad
d^{-}\left(i\right)=\min_{j\in\mathcal{P}_i^-}\left\|\hat{F}_d\left(i\right)-\hat{F}_d\left(j\right)\right\|_2^2.
\end{equation}
The $F_d$ is depth feature and $\hat{F}_d=F_d/\left\|F_d\right\|$. We aim to reduce the distance of the anchor from positive pixels while increasing the distance from negative pixels. However, if $d^{-}$ is too large, it will be detrimental to the training results, as the depth differences between neighboring objects are not always large. We set $s=0.65$ as a threshold to control the minimum distance between $d^{+}$ and $d^{-}$.
In addition, in the case of inaccurate prediction of the semantic segmentation model, we optimize to make both $|\mathcal{P}_i^+|$ and $|\mathcal{P}_i^-|$ larger than $r=4$:
\begin{equation}
d^+\left(i\right)+s<d^-\left(i\right),\mathrm{~}\forall\mathcal{P}_i\in\Lambda,
\end{equation}
\begin{equation}
\Lambda=\left\{\mathcal{P}_i\left|\begin{array}{c}(|\mathcal{P}_i^+|>r)\wedge(|\mathcal{P}_i^-|>r)\\\end{array}\right\}.\right. 
\end{equation}
The ${\Lambda}$ is the set of boundary pixels satisfying the appeal constraint. Then, the semantic boundary loss is defined as:
\begin{equation}
\mathcal{L}_{SBL}=\frac1{|\Lambda|}\sum_{\mathcal{P}_i\in\Lambda}\left(d^+(i)+\left[s-d^{-}(i)\right]_+\right),
\end{equation}
where the formula $[\cdot]_{+}$ is the hinge function. The above method enhances the model's semantic representation to cope with complex scenes.

\section{Experiments}

This part will first describe our experimental setup and evaluation metrics. The model's performance is then evaluated on both indoor and outdoor datasets. To verify the effectiveness of each design, we also conduct ablation experiments on the KITTI dataset.

\subsection{Experimental setup}

\subsubsection{Dataset}

\textbf{KITTI} \citep{kitti} contains 61 scenes captured by a moving vehicle equipped with multiple sensors, including the camera, LIDAR sensor, etc. For the fairness of evaluation, we use the data split of Eigen et al. \citeyearpar{Eigen}, which has 39,180 monocular triplets for training, 4,424 for evaluation, and 697 for testing.

\textbf{Make3D} \citep{make3d} is an outdoor dataset for depth estimation and scene understanding tasks. It can be used to test the generalization performance of the model. The dataset contains 134 outdoor images for evaluation. We train the model on the KITTI dataset and evaluate it directly on the Make3D dataset.

\textbf{NYU Depth V2} \citep{nyudepthv2} consists of various indoor video sequences captured with the RGB and depth cameras from the Microsoft Kinect. The dataset has 1,449 pairs of densely labeled RGB and depth images. Like Peng et al. \citeyearpar{EPCdepth}, we choose 654 images for evaluation and utilize the same metrics as the KITTI.

\subsubsection{Implementation details}
We implement the model on PyTorch and execute it on a single NVIDIA RTX2080Ti with 11GB of memory. The depth encoder adopts MPViT-Tiny \citep{mpvit}, while GhostNetV2-1.0 \citep{ghostnetv2} is the pose encoder. During training, the depth encoder and the pose encoder are pre-trained on ImageNet \citep{imagenet}, and then the model is trained on the KITTI dataset for 20 epochs. We use AdamW as the optimizer and set the initial learning rate for PoseNet and the depth decoder to $1\times10^{-4}$, and the initial learning rate of the depth encoder is $5\times10^{-5}$. The learning rate scheduler employs exponential decay with a decay rate of 0.9. In the comparison experiments, we set the resolution of the KITTI dataset to $640 \times 192$, and the range of evaluation depth is [0,80] meters. The resolution of the Make3D dataset is the same as the KITTI, and the predicted depth is adjusted to [0,70] meters. Moreover, the resolution of the NYU Depth V2 dataset is $384 \times 288$, and the range of evaluation depth is reduced to [0,10] meters. In the ablation experiments, the backbone versions we used are Swin Transformer adopted by SwinDepth \citep{Swindepth}, ResNet18 \citep{resnet}, MobileNetV3-Large-1.0 \citep{mobilenetv3}, GhostNetV2-1.0 \citep{ghostnetv2}, and MobileViT-Small \citep{mobilevit}, respectively. In addition, the weight of the semantic boundary loss is 0.1, and the channel reduction ratios of $\mathbf{Q}$ and $\mathbf{K}$ for CPA are set to 8.

\subsubsection{Metrics}
For evaluation, we employ a range of metrics proposed in \citep{metrics}. Absolute relative error (Abs Rel) measures the absolute difference between estimated and actual values relative to the actual value, directly comparing prediction accuracy. Squared relative error (Sq Rel) further emphasizes larger errors by squaring the differences. Root mean squared error (RMSE) calculates the square root of the mean of squared differences, providing a comprehensive measure of overall prediction error. Root mean squared logarithmic error (RMSE log) compares the logarithms of estimated and actual values and is suitable for a large range of values. Finally, threshold accuracy ($\delta$) measures the proportion of estimated values that match the actual values within a certain error threshold. The detailed metrics are as follows:
\begin{itemize} \item $\text{Abs Rel}=\frac1{||\mathcal{I}||}\sum_{p\in\mathcal{I}}\frac{|d(p)-d^*(p)|}{d^*(p)}.$ \item $\text{Sq Rel}=\frac1{||\mathcal{I}||}\sum_{p\in\mathcal{I}}\frac{(d(p)-d^*(p))^2}{d^*}.$ \item $\mathrm{RMSE}=\sqrt{\frac1{||\mathcal{I}||}\sum_{p\in\mathcal{I}}(d(p)-d^*(p))^2}.$\item $\text{RMSE log}=\sqrt{\frac1{||\mathcal{I}||}\sum_{p\in\mathcal{I}}(\log d(p)-\log d^*(p))^2}.$ \item $\delta=\frac1{||\mathcal{I}||}\sum_{p\in\mathcal{I}}\max\left(\frac d{d^*},\frac{d^*}d\right)<1.25^k.$
\end{itemize}
The $d$ and $d^{*}$ represent the estimated and actual depth values, respectively. The $p$ is a single pixel in the input image $\mathcal{I}$ and $\left|\left|\mathcal{I}\right|\right|$ is the total number of pixels in $\mathcal{I}$.

\subsection{Comparison with the state-of-the-art methods}

\subsubsection{KITTI}
\begin{table}[width=0.9\linewidth,cols=3,pos=!t]
\renewcommand\arraystretch{1.3}
\setlength{\tabcolsep}{4.8pt}
\caption{Results on the KITTI dataset adopting Eigen split \citep{Eigen}. All models are trained using monocular sequences with $640 \times 192$ resolution. The $\downarrow$ means that lower is better, $\uparrow$ is the opposite, and the best performance is shown in \textbf{bold}.}\label{tablekitti}
\begin{tabular*}{\tblwidth}{@{} l|cccc|ccc@{} }
\toprule
Method           & Abs Rel $\downarrow$ & Sq Rel $\downarrow$ & RMSE $\downarrow$  &RMSE log $\downarrow$ & $\delta$<1.25 $\uparrow$ &$\delta$<1.25$^{2}$ $\uparrow$ & $\delta$<1.25$^{3}$ $\uparrow$  \\ \midrule
Monodepth2 \citeyearpar{monodepth2}       & 0.115 & 0.903 & 4.863 & 0.193 & 0.877 & 0.959 & 0.981   \\
Johnston et al. \citeyearpar{johnston}         & 0.106 & 0.861 & 4.699 & 0.185 & 0.889 & 0.962 & 0.982  \\
SGDepth \citeyearpar{SGdepth}         & 0.117 & 0.907 & 4.844 & 0.196 & 0.875 & 0.958 & 0.980   \\
CADepth \citeyearpar{caddepth}          & 0.110 & 0.812 & 4.686 & 0.187 & 0.882 & 0.962 & 0.983    \\
MLDANet \citeyearpar{MLDA-Net}          & 0.110 & 0.824 & 4.632 & 0.187 & 0.883 & 0.961 & 0.982  \\
VC-Depth \citeyearpar{VC-Depth}         & 0.112 & 0.816 & 4.715 & 0.190 & 0.880 & 0.960 & 0.982      \\
Poggi et al. \citeyearpar{mono-uncertainty} & 0.111 & 0.863 & 4.756 & 0.188 & 0.881 & 0.961 & 0.982   \\
Guizilini et al. \citeyearpar{Guizilini}  & 0.117 & 0.854 & 4.714 & 0.191 & 0.873 & 0.963 & 0.981  \\
Insta-DM \citeyearpar{Leeetal}       & 0.112 & 0.777 & 4.772 & 0.191 & 0.872 & 0.959 & 0.982   \\
Patil et al.\citeyearpar{Patiletal}      & 0.111 & 0.821 & 4.650 & 0.187 & 0.883 & 0.961 & 0.982 \\
SAFENet \citeyearpar{SAFENet}          & 0.112 & 0.788 & 4.582 & 0.187 & 0.878 & 0.963 & 0.983  \\
G2S R50 \citeyearpar{G2S}          & 0.112 & 0.894 & 4.852 & 0.192 & 0.877 & 0.958 & 0.981   \\
Wang et al. \citeyearpar{Wangetal}       & 0.106 & 0.799 & 4.662 & 0.187 & 0.889 & 0.961 & 0.982       \\
R-MSFM3 \citeyearpar{r-msfm} & 0.114 & 0.815 & 4.712 & 0.193 & 0.876 & 0.959 & 0.981  \\
R-MSFM6 \citeyearpar{r-msfm} & 0.112 & 0.806 & 4.704 & 0.191 & 0.878 & 0.960 & 0.981    \\
HR-Depth \citeyearpar{hrdepth}         & 0.109 & 0.792 & 4.632 & 0.185 & 0.884 & 0.962 & 0.983       \\
FSRE-Depth \citeyearpar{fsre}       & 0.105 & 0.722 & 4.547 & 0.182 & 0.886 & 0.964 & 0.984   \\
ScaleNet \citeyearpar{ScaleInvariant}   & 0.109 & 0.779 & 4.641 & 0.186 & 0.883 & 0.962 & 0.982    \\
BRNet \citeyearpar{BRNet}            & 0.105 & 0.698 & 4.462 & 0.179 & 0.890 & 0.965 & 0.984        \\
DynaDepth \citeyearpar{DynaDepth}        & 0.108 & 0.761 & 4.608 & 0.187 & 0.883 & 0.962 & 0.982         \\
Lite-Mono \citeyearpar{lite-mono}        & 0.107 & 0.765 & 4.561 & 0.183 & 0.886 & 0.963 & 0.983     \\
MonoFormer \citeyearpar{monoformer}       & 0.104 & 0.846 & 4.580  & 0.183 & 0.891 & 0.962 & 0.982  \\
SwinDepth \citeyearpar{Swindepth}        & 0.106 & 0.739 & 4.510  & 0.182 & 0.890 & 0.964 & 0.984  \\
Ours             & \textbf{0.104} & \textbf{0.705} & \textbf{4.455} & \textbf{0.179} & \textbf{0.891} & \textbf{0.965} & \textbf{0.984}   \\ \bottomrule
\end{tabular*}
\end{table}

\begin{figure}[!t]
	\centering
	\begin{minipage}[t]{0.243\linewidth}
		\vspace{3pt}
		\centerline{\includegraphics[width=\textwidth]{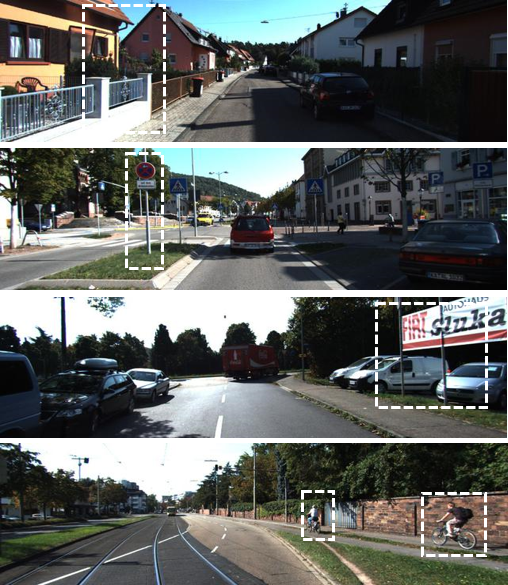}}
		\centerline{Input}
	\end{minipage}
	\begin{minipage}[t]{0.243\linewidth}
		\vspace{3pt}
		\centerline{\includegraphics[width=\textwidth]{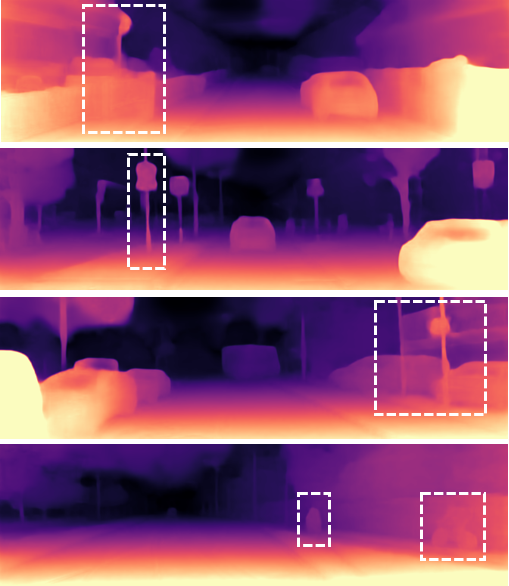}}
		\centerline{Monodepth2}
	\end{minipage}
	\begin{minipage}[t]{0.243\linewidth}
		\vspace{3pt}
		\centerline{\includegraphics[width=\textwidth]{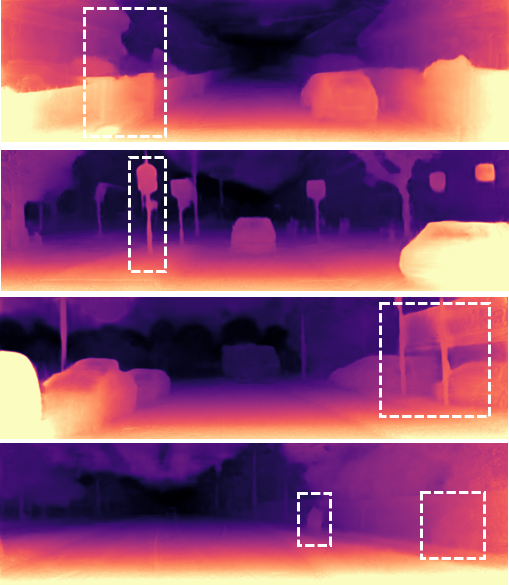}}
		\centerline{SwinDepth}
	\end{minipage}
	\begin{minipage}[t]{0.243\linewidth} 
		\vspace{3pt}
		\centerline{\includegraphics[width=\textwidth]{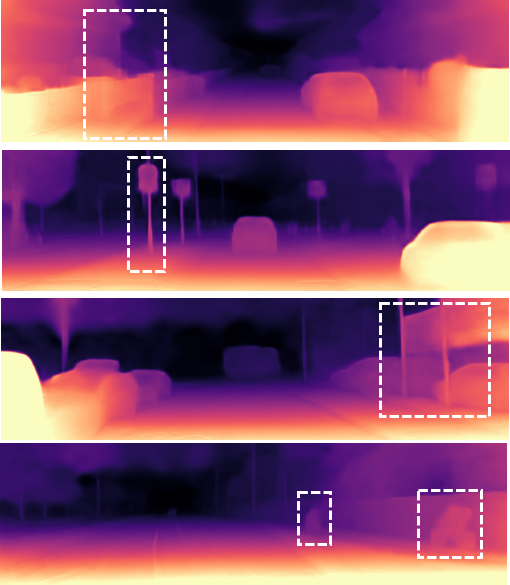}}
		\centerline{Ours}
	\end{minipage}
	\caption{Qualitative results on KITTI. Predictions by Monodepth2 \citep{monodepth2}, SwinDepth \citep{Swindepth} and ours. The proposed model can clearly show the cyclists that are almost ignored in SwinDepth. Meanwhile, the outline of roadside billboards and buildings is more complete.}
	\label{Fig .8}
\end{figure}

We train and evaluate the model on the KITTI dataset. For the fairness of the experiment, the compared models are trained using monocular image sequences with the same resolution. As illustrated in Table \ref{tablekitti}, benefiting from the guidance of multiple priors, our model achieves the best results on the seven evaluation metrics, among which the $\text{Abs Rel}$ metric achieves 0.104. To further demonstrate the representation of the proposed model, we conduct a qualitative comparison with other methods. Figure \ref{Fig .8} shows that our model has a more detailed representation capability and predicts the contours of cyclists, roadside billboards, and buildings clearly.

\subsubsection{Make3D}
\begin{figure}[!t]
	\centering
	\begin{minipage}[t]{0.235\linewidth}
		\vspace{3pt}
		\centerline{\includegraphics[width=\textwidth]{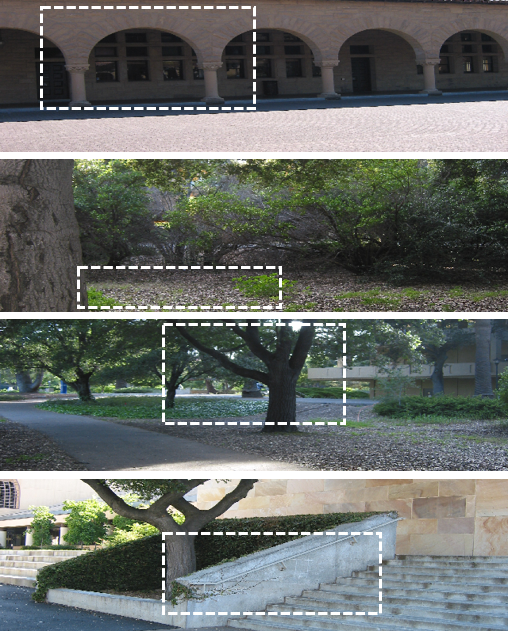}}
		\centerline{Input}
	\end{minipage}
	\begin{minipage}[t]{0.235\linewidth}
		\vspace{3pt}
		\centerline{\includegraphics[width=\textwidth]{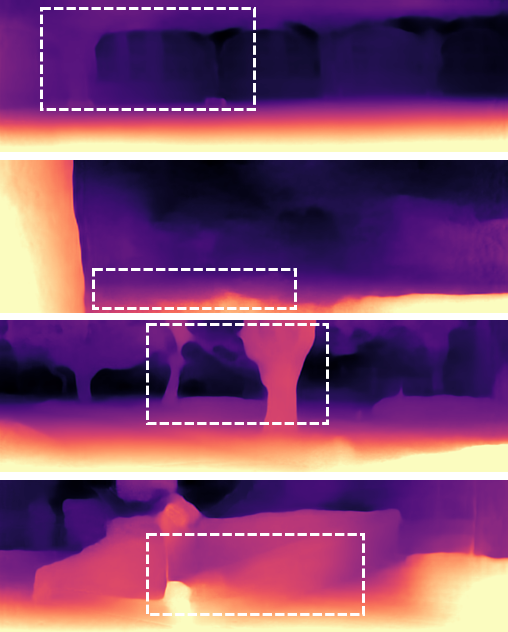}}
		\centerline{Monodepth2}
	\end{minipage}
	\begin{minipage}[t]{0.235\linewidth}
		\vspace{3pt}
		\centerline{\includegraphics[width=\textwidth]{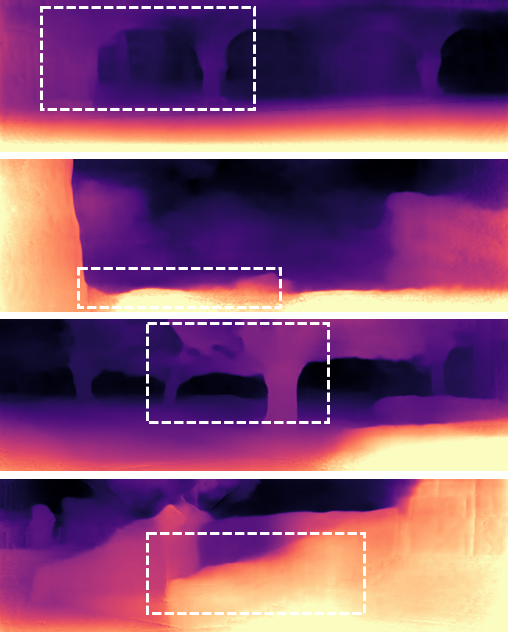}}
		\centerline{DynaDepth}
	\end{minipage}
	\begin{minipage}[t]{0.235\linewidth} 
		\vspace{3pt}
		\centerline{\includegraphics[width=\textwidth]{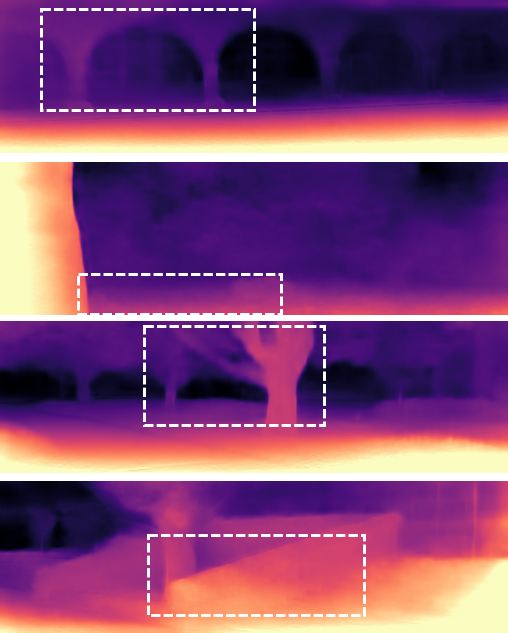}}
		\centerline{Ours}
	\end{minipage}
	\caption{Qualitative results on Make3D. Predictions by Monodepth2 \citep{monodepth2}, DynaDepth \citep{DynaDepth} and ours. The proposed model shows the complete outline of the buildings while being slightly affected by sparse leaves.}
	\label{FIG make3d}
\end{figure}

To verify the generalization ability, we evaluate the model on the Make3D dataset. Generally, the trained model gives better results when evaluated on a dataset similar to the training scenes. To avoid this, we adopt a different outdoor dataset for evaluation.
The model loads the weights trained on the KITTI dataset directly without fine-tuning. We follow the metrics of monodepth \citep{monodepth}, as shown in Table \ref{Table make3d}, and our model outperforms the state-of-the-art methods in six metrics. Meanwhile, we present qualitative results in Fig. \ref{FIG make3d}, where our model exhibits more complete contours of buildings and tree trunks, slightly affected by sparse leaves. The results prove that the variation of the outdoor scene does not affect the performance of our model.

\begin{table}[width=.9\linewidth,cols=3,pos=t]
\renewcommand\arraystretch{1.3}
\caption{Results on the Make3D dataset. The models are loaded with weights trained on the KITTI dataset with $640 \times 192$ resolution. The $\downarrow$ means that lower is better, $\uparrow$ is the opposite, and the best performance is shown in \textbf{bold}.}\label{Table make3d}
\begin{tabular*}{\tblwidth}{@{} L|CCCC|CCC@{} }
\toprule
Method & Abs Rel $\downarrow$ & Sq Rel $\downarrow$& RMSE $\downarrow$& RMSE log $\downarrow$& $\delta$<1.25 $\uparrow$& $\delta$<1.25$^{2}$ $\uparrow$& $\delta$<1.25$^{3}$ $\uparrow$\\
\midrule
Monodepth2 \citeyearpar{monodepth2} & 0.321 & 3.378 & 7.252 & 0.163 & 0.553 & 0.798 & 0.906 \\
FSRE-Depth \citeyearpar{fsre} & 0.325 & 3.238 & 7.189 & 0.166 & 0.538 & 0.784 & 0.902 \\
Lite-Mono \citeyearpar{lite-mono} & 0.305 & 3.060 & 6.981 & 0.158 & \textbf{0.572} & 0.806 & 0.911 \\
DynaDepth \citeyearpar{DynaDepth} & 0.316 & 3.171 & 7.060 & 0.162 & 0.551 & 0.796 & 0.909 \\
Ours & \textbf{0.304} & \textbf{2.952} & \textbf{6.859} & \textbf{0.156} & 0.565 & \textbf{0.812} & \textbf{0.917} \\
\bottomrule
\end{tabular*}
\end{table}

\subsubsection{NYU Depth V2}
\begin{table}[width=.9\linewidth,cols=3,pos=t]
\renewcommand\arraystretch{1.3}
\caption{Results on the NYU Depth V2 dataset. The models are loaded with weights trained on the KITTI dataset with $640 \times 192$ resolution. The $\downarrow$ means that lower is better, $\uparrow$ is the opposite, and the best performance is shown in \textbf{bold}.}\label{Table nyu}
\begin{tabular*}{\tblwidth}{@{} L|CCCC|CCC@{} }
\toprule
Method & Abs Rel $\downarrow$ & Sq Rel $\downarrow$& RMSE $\downarrow$& RMSE log $\downarrow$& $\delta$<1.25 $\uparrow$& $\delta$<1.25$^{2}$ $\uparrow$& $\delta$<1.25$^{3}$ $\uparrow$\\
\midrule
Monodepth2 \citeyearpar{monodepth2} & 0.355 & 0.673 & 1.252 & 0.373 & 0.485 & 0.771 & 0.907 \\
DynaDepth \citeyearpar{DynaDepth} & 0.335 & 0.589 & 1.189 & 0.361 & 0.502 & 0.784 & 0.916 \\
Lite-Mono \citeyearpar{lite-mono} & 0.329 & 0.546 & 1.169 & 0.363 & 0.499 & 0.781 & 0.915 \\
FSRE-Depth \citeyearpar{fsre} & 0.330 & 0.581 & 1.211 & 0.359 & 0.510 & 0.790 & 0.918 \\
Ours & \textbf{0.308} & \textbf{0.523} & \textbf{1.160} & \textbf{0.343} & \textbf{0.530} & \textbf{0.808} & \textbf{0.929} \\
\bottomrule
\end{tabular*}
\end{table}

\begin{figure}[!t]
	\centering
	\begin{minipage}[t]{0.15\linewidth}
		\vspace{3pt}
		\centerline{\includegraphics[width=\textwidth]{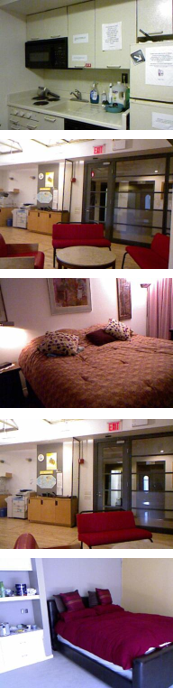}}
		\centerline{Input}
	\end{minipage}
	\begin{minipage}[t]{0.15\linewidth}
		\vspace{3pt}
		\centerline{\includegraphics[width=\textwidth]{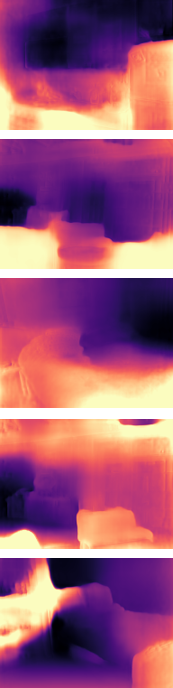}}
		\centerline{Monodepth2}
	\end{minipage}
	\begin{minipage}[t]{0.15\linewidth}
		\vspace{3pt}
		\centerline{\includegraphics[width=\textwidth]{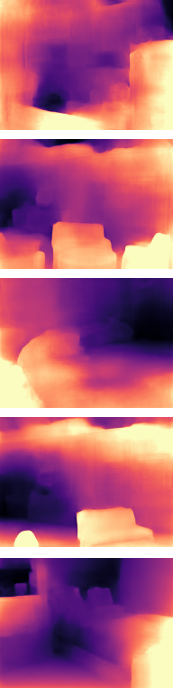}}
		\centerline{FSRE-Depth}
	\end{minipage}
	\begin{minipage}[t]{0.15\linewidth}
		\vspace{3pt}
		\centerline{\includegraphics[width=\textwidth]{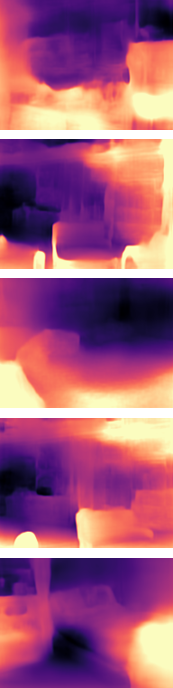}}
		\centerline{Lite-Mono}
	\end{minipage}
	\begin{minipage}[t]{0.15\linewidth}
		\vspace{3pt}
		\centerline{\includegraphics[width=\textwidth]{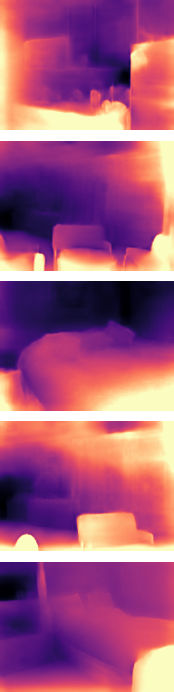}}
		\centerline{Ours}
	\end{minipage}
	\begin{minipage}[t]{0.15\linewidth}
		\vspace{3pt}
		\centerline{\includegraphics[width=\textwidth]{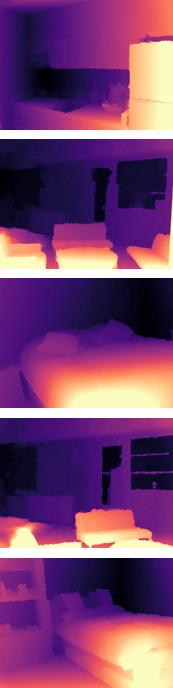}}
		\centerline{GT}
	\end{minipage}
	\caption{Qualitative results on NYU Depth V2. Predictions by Monodepth2 \citep{monodepth2}, FSRE-Depth \citep{fsre}, Lite-Mono \citep{lite-mono} and ours. The GT represents the ground truth. The predictions of our model are the closest to the ground truth, and the indoor furniture, appliances, and even pillows on the bed can be estimated clearly.}
	\label{Fig .nyu}
\end{figure}
Depth estimation models trained on outdoor datasets are better at handling objects such as vehicles, pedestrians, and buildings, while the details of indoor scenes are difficult to estimate. To further verify the generalization performance, we transfer the scenes from outdoor to indoor. Our model and previous works use the weights trained on the KITTI dataset and then evaluated on the NYU Depth V2 dataset.
As shown in Table \ref{Table nyu}, on the important $\text{Abs Rel}$ metric, we achieve 0.308, exceeding the recent method \citep{lite-mono} by 0.021. Meanwhile, we present the qualitative results of our method versus other methods in Fig. \ref{Fig .nyu}, where our model generates higher quality depth maps that can estimate furniture, appliances, and even pillows on a bed.
Through the experiment, it can be proved that even if the model is trained with the outdoor dataset, it can also be applied well in indoor scenes.

\subsubsection{Complexity analysis}
We analyze the complexity of the model. While comparing its accuracy, we also evaluate its floating point operations per second (FLOPs), GPU memory usage (VRAM), params, model size, and training devices. In Table \ref{Table complexity}, our model achieves optimal accuracy while having lower FLOPs, VRAM, and params. The FLOPs is 11.9G, representing a decrease of an order of magnitude compared to the method that utilizes the transformer architecture \citep{Swindepth}. Unlike models requiring a Tesla V100 or multiple GPUs, our model completes training on a single RTX 2080Ti. This advantage reduces hardware costs, enabling more researchers and developers to train and utilize the model with limited resources.

\begin{table}[width=.9\linewidth,cols=3,pos=t]
\renewcommand\arraystretch{1.3}
\setlength{\tabcolsep}{3.5pt}
\caption{Complexity analysis on the KITTI dataset. We compare models in floating point operations per second (FLOPs), GPU memory usage (VRAM), params, model size, and training devices. The best performance is shown in \textbf{bold}.}\label{Table complexity}
\begin{tabular*}{\tblwidth}{@{} L|CCCCCC@{} }
\toprule
Method    & Abs Rel & FLOPs       & VRAM    & Params  & Model size & Device \\
\midrule
BRNet \citeyearpar{BRNet}     & 0.105   & 33.7 G  & 1941 MiB & 19.12 M & 79.7 MB     & V100   \\
HR-Depth \citeyearpar{hrdepth}  & 0.109   & 31.6 G  & 1769 MiB & 16.96 M & 58.6 MB     & V100   \\
DynaDepth \citeyearpar{DynaDepth} & 0.108   & 16.6 G  & 2261 MiB & 32.52 M & 138.6 MB    & V100   \\
SwinDepth \citeyearpar{Swindepth}& 0.106   & 132.9 G & 2361 MiB & 25.13 M & 100.8 MB    & 4 $\times$ 3090 \\
Ours      & \textbf{0.104}   & \textbf{11.9} G  & \textbf{1851} MiB & \textbf{8.43} M  & \textbf{33.2} MB     & 2080Ti\\
\bottomrule
\end{tabular*}
\end{table}
\subsection{Ablation studies}
To prove the rationality of the model, we conduct ablation experiments on the KITTI dataset. First, we evaluate the impact of different backbones on depth estimation. Then, the rationality of the CPA is verified by adjusting the structure of the channel branch and the channels of the spatial branch. The model consists of multiple losses, and different loss ratios affect the performance. To find an appropriate loss ratio, we set different loss ratios according to the gradient for training. Finally, the ablation experiment is performed on the whole framework of the model.

\subsubsection{Encoder}

\begin{figure}[!t]
	\centering
	\begin{minipage}[t]{0.24\linewidth}
		\vspace{3pt}
		\centerline{\includegraphics[width=\textwidth]{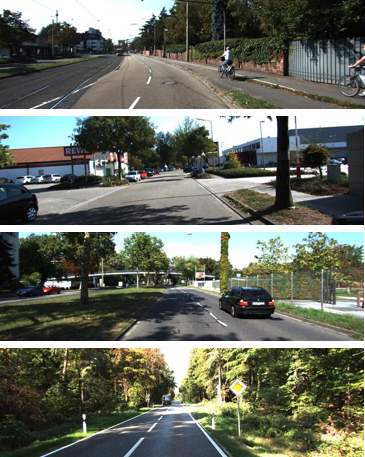}}
		\centerline{Input}
	\end{minipage}
	\begin{minipage}[t]{0.24\linewidth}
		\vspace{3pt}
		\centerline{\includegraphics[width=\textwidth]{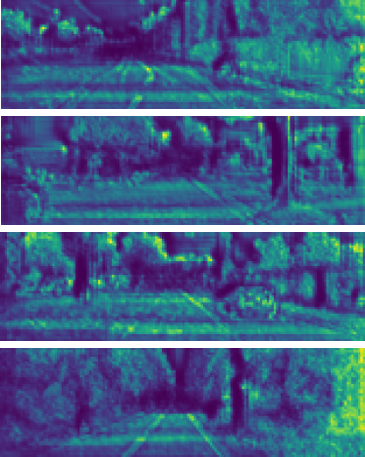}}
		\centerline{ResNet18}
	\end{minipage}
	\begin{minipage}[t]{0.24\linewidth}
		\vspace{3pt}
		\centerline{\includegraphics[width=\textwidth]{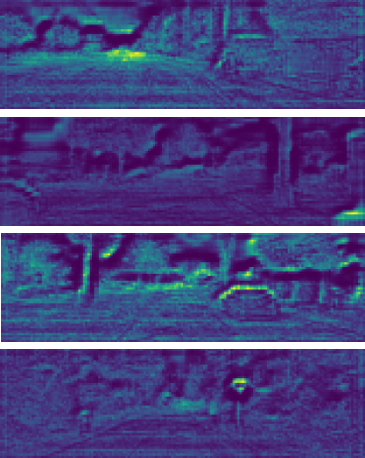}}
		\centerline{Swin Transformer}
	\end{minipage}
	\begin{minipage}[t]{0.24\linewidth}
		\vspace{3pt}
		\centerline{\includegraphics[width=\textwidth]{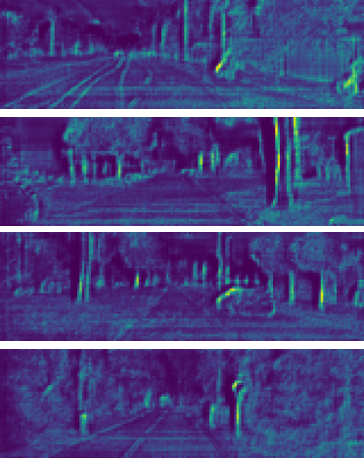}}
		\centerline{Ours}
	\end{minipage}
	\caption{Features of the depth backbones on KITTI. The features generated by ResNet18 \citep{resnet}, Swin Transformer \citep{Swindepth}, and ours. Compared with ResNet18, the transformer architectures are more prominent in spatial structure representation. Meanwhile, our model can highlight road signs, cyclists, and vehicles compared to the Swin Transformer.}
	\label{Fig .backbone}
\end{figure}

For the encoder ablation experiment, the depth decoder and PoseNet follow Monodepth2 \citep{monodepth2}, and the initial learning rate of the network is $1\times10^{-4}$.
Subsequently, we employ a series of CNNs and transformers as depth backbones for training. Due to the RTX 2080Ti not meeting the memory requirements of Swin Transformer \citep{Swindepth}, we additionally train the model with Swin Transformer on RTX 3090Ti. The results are shown in Table \ref{Table backbone}, the transformer structures have higher accuracy than CNNs. Moreover, our model outperforms MobileViT \citep{mobilevit} and Swin Transformer in $\text{Abs Rel}$ metric reaching 0.110. At the same time, our memory usage is less than half of the Swin Transformer. Further, we qualitatively present the features of the backbones in Fig. \ref{Fig .backbone}. The transformer architectures exhibit spatial structure more obvious than ResNet18, while our model highlights objects in the scene, such as road signs, cyclists, and vehicles compared to the Swin Transformer.

\begin{table}[width=.9\linewidth,cols=3,pos=!t]
\renewcommand\arraystretch{1.3}
\setlength{\tabcolsep}{3.5pt}
\caption{Results for the depth backbones on the KITTI dataset. We list the GPU memory usage (VRAM) during training. The input resolution of the model is $640 \times 192$. The $\downarrow$ means that lower is better, $\uparrow$ is the opposite, and the best performance is shown in \textbf{bold}.}\label{Table backbone}
\begin{tabular*}{\tblwidth}{@{} L|CCCC|CCC|C@{} }
\toprule
Method & Abs Rel $\downarrow$ & Sq Rel $\downarrow$ & RMSE $\downarrow$ & RMSE log $\downarrow$ & $\delta$<1.25 $\uparrow$ & $\delta$<1.25$^{2}$ $\uparrow$ & $\delta$<1.25$^{3}$ $\uparrow$ & VRAM\\
\midrule
ResNet18 \citeyearpar{resnet} & 0.115 & 0.903 & 4.863 & 0.193 & 0.877 & 0.959 & 0.981 & \textbf{5076}\\
MobileNetV3 \citeyearpar{mobilenetv3} & 0.119 & 0.920 & 4.891 & 0.196 & 0.871 & 0.958 & 0.980 & 6064 \\
GhostNetV2 \citeyearpar{ghostnetv2} & 0.119 & 0.933 & 4.920 & 0.195 & 0.867 & 0.958 & 0.981 & 6056\\
MobileViT \citeyearpar{mobilevit}  & 0.112 & 0.833 & 4.645 & 0.188 & 0.883 & 0.962 & 0.982 & 7014 \\
Swin Transformer \citeyearpar{Swindepth} & 0.113 & \textbf{0.775} & \textbf{4.619} & 0.187 & 0.878 & \textbf{0.963} & \textbf{0.983} & 15798 \\
Ours & \textbf{0.110} & 0.836 & 4.646 & \textbf{0.186} & \textbf{0.886} & 0.962 & 0.982 & 6602 \\
\bottomrule
\end{tabular*}
\end{table}

\subsubsection{Context prior attention}

The CPA combines the channel and spatial branches to complement the context representation. To prove the rationality of the structure, we also combine other channel attention mechanisms with the spatial branch in a parallel manner. Table \ref{Table cpa} shows the comparison to other models, and our model achieves the best results on five metrics. Furthermore, in the spatial branch, the channel reduction ratios of $\mathbf{Q}$ and $\mathbf{K}$ affect the number of parameters in CPA. The parameters decrease as the channel reduction ratio increases, but the accuracy also fluctuates. To adjust appropriate channels for CPA. The reduction ratios of $\mathbf{Q}$ and $\mathbf{K}$ are set to $4$, $6$, $8$, $12$, and $16$, respectively. Through the experiment shown in Table \ref{Table cpac}, the model achieves the best performance when the channel reduction ratio is 8, and the $\text{Abs Rel}$ metric achieves 0.104.

\begin{table}[width=.9\linewidth,cols=3,pos=t]
\renewcommand\arraystretch{1.3}
\caption{Results on the structure of CPA. The $^{*}$ represents the channel attention of ECANet \citep{ecanet}, DANet \citep{danet}, SENet \citep{senet}, and CBAM \citep{cbam} combined with the spatial branch of CPA, respectively. The pose encoder adopts ResNet18. The $\downarrow$ means that lower is better, $\uparrow$ is the opposite, and the best performance is shown in \textbf{bold}.}\label{Table cpa}
\begin{tabular*}{\tblwidth}{@{} L|CCCC|CCC@{} }
\toprule
Method & Abs Rel $\downarrow$ & Sq Rel $\downarrow$ & RMSE $\downarrow$ & RMSE log $\downarrow$ & $\delta$<1.25 $\uparrow$ & $\delta$<1.25$^{2}$ $\uparrow$ & $\delta$<1.25$^{3}$ $\uparrow$\\
\midrule
CCNet \citeyearpar{ccnet} & 0.104 & \textbf{0.707} & \textbf{4.480} & 0.178 & 0.891 & 0.964 & 0.984\\
ECANet$^{*}$ \citeyearpar{ecanet} & 0.107 & 0.757 & 4.532 & 0.181 & 0.887 & 0.964 & 0.984\\
DANet$^{*}$ \citeyearpar{danet} & 0.107 & 0.760 & 4.506 & 0.180 & 0.890 & 0.965 & 0.984\\
SENet$^{*}$ \citeyearpar{senet}  & 0.108 & 0.760 & 4.513 & 0.179 & 0.888 & 0.965 & 0.984\\
CBAM$^{*}$ \citeyearpar{cbam} & 0.107 & 0.768 & 4.545 & 0.179 & 0.891 & 0.965 & 0.984\\
Ours & \textbf{0.104} & 0.751 & 4.494 & \textbf{0.177} & \textbf{0.893} & \textbf{0.965} & \textbf{0.984}\\
\bottomrule
\end{tabular*}
\end{table}

\begin{table}[width=.9\linewidth,cols=3,pos=t]
\renewcommand\arraystretch{1.3}
\caption{Results on the channel of CPA. We explore the effect of the channel reduction ratios of $\mathbf{Q}$ and $\mathbf{K}$ on the spatial branch. The channel reduction ratio of our model is 8. Meanwhile, the pose encoder adopts ResNet18. The $\downarrow$ means that lower is better, $\uparrow$ is the opposite, and the best performance is shown in \textbf{bold}.}\label{Table cpac}
\begin{tabular*}{\tblwidth}{@{} C|CCCC|CCC@{} }
\toprule
Reduction Ratio & Abs Rel $\downarrow$ & Sq Rel $\downarrow$ & RMSE $\downarrow$ & RMSE log $\downarrow$ & $\delta$<1.25 $\uparrow$ & $\delta$<1.25$^{2}$ $\uparrow$ & $\delta$<1.25$^{3}$ $\uparrow$\\
\midrule
4 & 0.105 & 0.748 & 4.488 & 0.179 & 0.893 & 0.965 & 0.984\\
6 & 0.105 & \textbf{0.707} & \textbf{4.443} & 0.178 & 0.889 & 0.965 & 0.984\\
12 & 0.105 & 0.736 & 4.446 & 0.178 & 0.892 & 0.965 & 0.984\\
16 & 0.104 & 0.736 & 4.496 & 0.178 & 0.890 & 0.965 & 0.984\\
Ours & \textbf{0.104} & 0.751 & 4.494 & \textbf{0.177} & \textbf{0.893} & \textbf{0.965} & \textbf{0.984}\\
\bottomrule
\end{tabular*}
\end{table}

\subsubsection{Semantic boundary loss}
The loss function determines how the model optimizes its parameters during training. In our model, the loss function comprises reprojection loss \citep{monodepth2}, smoothness loss \citep{monodepth}, and semantic boundary loss. We conduct experiments to investigate the impact of different loss ratios on the model's performance.
As shown in Table \ref{Table sbl}, we first attempt to use only reprojection and smoothness loss, with a weight ratio of 1:1, achieving an Abs Rel metric of 0.108. We introduce semantic boundary loss to enhance semantic representation, initially setting its weight to 1. As the weight of the semantic boundary loss gradually decreased, we observed a significant improvement in the evaluation metrics. Specifically, when the weight of the semantic boundary loss is set to 0.1, the model achieves its optimal performance.
This indicates that when the ratio between reprojection loss, smoothness loss, and semantic boundary loss is 1:1:0.1, the model is able to balance the optimization needs of various aspects best, thereby achieving optimal performance.

\begin{table}[width=.93\linewidth,cols=3,pos=t]
\renewcommand\arraystretch{1.3}
\caption{Results on the loss ratio. We test the effect of different loss ratios on the KITTI dataset. The RL, SL, and SBL represent reprojection loss, smoothness loss, and semantic boundary loss, respectively. Our model's loss ratio is 1:1:0.1. The $\downarrow$ means that lower is better, $\uparrow$ is the opposite, and the best performance is shown in \textbf{bold}.}\label{Table sbl}
\begin{tabular*}{\tblwidth}{@{} CCC|CCCC|CCC@{} }
\toprule
RL & SL & SBL & Abs Rel $\downarrow$ & Sq Rel $\downarrow$ & RMSE $\downarrow$ & RMSE log $\downarrow$ & $\delta$<1.25 $\uparrow$ & $\delta$<1.25$^{2}$ $\uparrow$ & $\delta$<1.25$^{3}$ $\uparrow$\\
\midrule
1 & 1 & \textemdash & 0.108 & 0.757 & 4.541 & 0.181 & 0.886 & 0.964 & 0.984 \\
1 & 1 & 1 & 0.107 & 0.736 & 4.464 & 0.178 & 0.885 & 0.965 & 0.985 \\
1 & 1 & 0.8 & 0.107 & 0.729 & 4.520 & 0.179 & 0.884 & 0.964 & 0.985\\
1 & 1 & 0.5 & 0.105 & \textbf{0.702} & 4.456 & 0.177 & 0.888 & 0.966 & \textbf{0.985} \\
1 & 1 & 0.2 & 0.105 & 0.722 & 4.470 & \textbf{0.177} & 0.887 & \textbf{0.966} & 0.985 \\
1 & 1 & 0.1 & \textbf{0.104} & 0.705 & \textbf{4.455} & 0.179 & \textbf{0.891} & 0.965 & 0.984 \\
\bottomrule
\end{tabular*}
\end{table}

\subsubsection{Framework}
\begin{figure}[!t]
	\centering
	\begin{minipage}[t]{0.24\linewidth}
		\vspace{3pt}
		\centerline{\includegraphics[width=\textwidth]{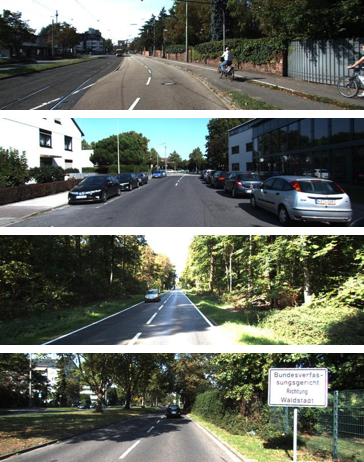}}
		\centerline{Input}
	\end{minipage}
	\begin{minipage}[t]{0.24\linewidth}
		\vspace{3pt}
		\centerline{\includegraphics[width=\textwidth]{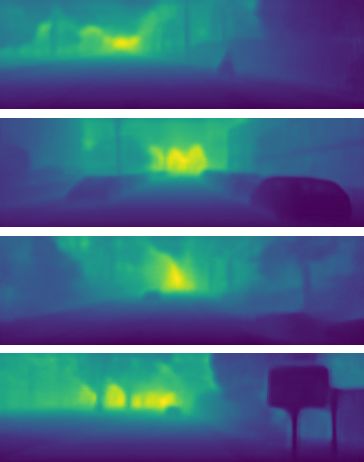}}
		\centerline{W/O SBL}
	\end{minipage}
	\begin{minipage}[t]{0.24\linewidth}
		\vspace{3pt}
		\centerline{\includegraphics[width=\textwidth]{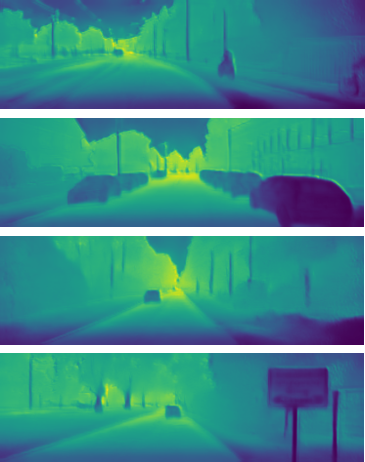}}
		\centerline{SBL}
	\end{minipage}
	\begin{minipage}[t]{0.24\linewidth}
		\vspace{3pt}
		\centerline{\includegraphics[width=\textwidth]{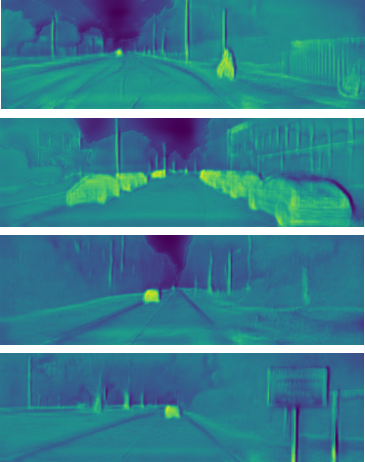}}
		\centerline{SBL $+$ SPA}
	\end{minipage}
	\caption{Visualization of semantic prior analysis. We list the cases without semantic boundary loss, semantic boundary loss, and semantic boundary loss $+$ semantic prior attention. After introducing semantic boundary loss, the scene contour representation is clearer. Further semantic prior attention highlights objects on the street.}
	\label{Fig. sbl}
\end{figure}

To verify the effectiveness of the proposed modules, we conduct the ablation experiment on the whole framework. As shown in Table \ref{Table framework}, based on our baseline \citep{monodepth2}, we replace the backbone with MPViT-Tiny. Benefiting from the transformer architecture that can focus on spatial information, the $\text{Abs Rel}$ metric is improved by 0.009. We introduce the SBL into the model, with four metrics improved and two decreased. Meanwhile, the promotion is limited when only the SPA is used without SBL. However, when both SBL and SPA are used in the model, the $\text{Abs Rel}$ metric is maintained, while four metrics are improved. We analyze that the introduction of SBL enables the model to learn the contour information of the scene but does not focus on the objects on the street. The use of SPA alleviates this problem. To further prove this point, we visualize the feature map of the decoder in Fig. \ref{Fig. sbl}, where the introduction of SBL makes the contour of the scene clearer. Additionally, using SPA highlights objects on the street, such as vehicles and cyclists. Subsequently, the CPA improves five metrics, of which the $\text{Abs Rel}$ metric is improved by 0.002. Finally, we use the lightweight GhostNetV2 as the pose backbone to extract spatial pose priors and achieve the best performance.

\begin{table}[width=.93\linewidth,cols=3,pos=!t]
\renewcommand\arraystretch{1.3}
\setlength{\tabcolsep}{2.0pt}
\caption{Results of the ablation experiment for the overall structure. MPViT: MPViT-Tiny. SBL: semantic boundary loss. SPA: semantic prior attention. CPA: context prior attention. GhostNetV2: lightweight pose backbone. The $\downarrow$ means that lower is better, $\uparrow$ is the opposite, and the best performance is shown in \textbf{bold}.}\label{Table framework}
\begin{tabular*}{\tblwidth}{@{} ccccc|cccc|ccc@{} }
\toprule
 MPViT & SBL & SPA & CPA & GhostNetV2 & Abs Rel $\downarrow$ & Sq Rel $\downarrow$ & RMSE $\downarrow$ & RMSE log $\downarrow$ & $\delta$<1.25 $\uparrow$ & $\delta$<1.25$^{2}$ $\uparrow$ & $\delta$<1.25$^{3}$ $\uparrow$\\ \midrule
       &     &        &     &            & 0.115   & 0.903  & 4.863 & 0.193    & 0.877 & 0.959 & 0.981 \\
 \checkmark     &     &        &     &            & 0.106   & 0.737  & 4.486 & 0.182    & 0.889 & 0.963 & 0.983 \\
 \checkmark     & \checkmark   &        &     &            & 0.107   & 0.750  & 4.482 & 0.181    & 0.890  & 0.965 & 0.983 \\
 \checkmark     &     & \checkmark      &     &            & 0.107   & 0.788  & 4.570  & 0.181    & 0.890  & 0.964 & 0.983 \\
 \checkmark     & \checkmark   & \checkmark      &     &            & 0.106   & 0.745  & 4.516 & 0.181    & 0.890  & 0.964 & 0.984 \\
 \checkmark     & \checkmark   & \checkmark      & \checkmark   &            & 0.104   & 0.751  & 4.494 & \textbf{0.177}    & \textbf{0.893} & 0.965 & 0.984 \\
 \checkmark     & \checkmark   & \checkmark      & \checkmark   & \checkmark          & \textbf{0.104}   & \textbf{0.705}  & \textbf{4.455} & 0.179    & 0.891 & \textbf{0.965} & \textbf{0.984} \\ \bottomrule
\end{tabular*}
\end{table}

\subsection{Limitation and discussion}
\subsubsection{Limitation}
While the work has achieved certain outcomes, it still has some limitations. The proposed semantic prior method relies on a pre-trained semantic segmentation network to generate semantic pseudo-labels, which adds complexity to the entire training process and may limit the accuracy of depth estimation due to the performance bottleneck of the semantic segmentation network. To reduce the dependence on external networks and simplify the training process, we can consider exploring more efficient ways to acquire semantic information, such as using self-supervised methods to learn semantic features directly. On the other hand, while self-supervised learning avoids the expensive cost of manual annotations, the vast amount of training data still makes the training process time-consuming. For instance, training on the KITTI dataset using an RTX 2080Ti takes over 16 hours, increasing research time and potentially limiting the iteration and optimization speed of the algorithm. Future research can explore more effective optimization algorithms, data augmentation techniques, or hardware acceleration solutions to improve training efficiency and reduce training time.
In addition, self-supervised monocular depth estimation often suffers from a limited depth prediction range. The algorithm may not provide accurate depth estimates for objects or scenes beyond this range, affecting the stability in ultra-long-range views. To address this issue, we can explore multi-scale depth estimation methods to handle scenes with different distance ranges better.
\subsubsection{Discussion}
The semantic priors enhance the representation of the model while also leading to additional data preparation. As shown in Fig. \ref{Fig .backbone}, the features extracted from the backbone initially highlighted the outline of the objects. To further refine these representations, we can consider using the attention mechanism to weigh contours in the scene and thus focus on object boundaries. Furthermore, we can leverage multi-task learning \citep{multitask} to train depth estimation and other tasks simultaneously. In this way, the model can utilize the semantic information of the semantic segmentation task to assist in depth estimation and refine the contour representation. To address the training efficiency challenge, knowledge distillation \citep{distillation} emerges as a promising solution. It facilitates knowledge transfer from a larger, pre-trained model (teacher model) to a smaller model (student model), significantly accelerating the training process of the latter. Since the teacher model is trained on vast amounts of data, its guidance tends to be more effective than training on raw data alone. Moreover, to address the uncertainty and fuzziness of information, we can incorporate techniques such as the q-rung orthopair fuzzy algorithm \citep{q-rung}, fuzzy multi-criteria modeling \citep{multi-criteria}, and fuzzy clustering feature extraction \citep{clustering}. These methods provide more robust ways to handle imprecise and ambiguous information. Lastly, we note that few-shot learning \citep{fewshot} has recently gained significant attention in computer vision. Combining few-shot learning with self-supervised monocular depth estimation can reduce the amount of training data. This direction holds immense promise for low-cost training.
\section{Conclusion}
This paper designs a novel self-supervised monocular depth estimation model that overcomes the limitation of relying on labeled information by integrating spatial, context, and semantic prior. We utilize a hybrid transformer and lightweight pose network to learn long-range spatial prior, enhancing the model's understanding of global spatial relationships. The designed context prior attention mechanism significantly improves the model's generalization capability in complex structures and untextured regions. Additionally, we introduce semantic boundary loss and semantic prior attention mechanism to guide the model in learning semantic prior information and address the issue of boundary scale deviation. Experimental results demonstrate that the model performs excellently on the KITTI, Make3D, and NYU Depth V2 datasets, verifying its effectiveness and robustness.

Future work will focus on the following directions: we will further explore more types of prior information and investigate how to effectively fuse them to enhance the model's representation ability. Then, focus on the robustness under conditions such as nighttime and severe weather to adapt the model to these complex environments. Moreover, exploring scene understanding based on multi-task learning and expanding the model to fields such as robot navigation and industrial applications.

\bibliographystyle{plainnat}
\bibliography{refs}
\end{document}